\documentclass[conference]{IEEEtran}
\usepackage{times}
\usepackage{float}
\usepackage{wrapfig}
\usepackage{tikz}
\usepackage[skip=0pt]{caption}
\usepackage{subcaption}
\usepackage{multicol}
\usepackage{footmisc}
\usepackage{xifthen}
\usepackage{algorithm}
\usepackage{algpseudocode}
\usepackage{xcolor}
\usepackage{graphicx}
\usepackage{dblfloatfix}
\usepackage{afterpage}
\usepackage[numbers]{natbib}
\usepackage{multicol}
\usepackage[bookmarks=true]{hyperref}
\usepackage[utf8]{inputenc}

\usepackage{amsthm}
\theoremstyle{definition}
\newtheorem{definition}{Definition}

\usepackage{amsmath,amsfonts,bm}

\def\eqref#1{equation~\ref{#1}}

\def\1{\bm{1}}

\def\rx{{\textnormal{x}}}
\def\ry{{\textnormal{y}}}

\def\rvk{{\mathbf{k}}}

\def\rvp{{\mathbf{p}}}

\def\rvz{{\mathbf{z}}}

\def\vc{{\bm{c}}}

\def\vf{{\bm{f}}}

\def\vh{{\bm{h}}}

\def\vl{{\bm{l}}}

\def\vp{{\bm{p}}}
\def\vq{{\bm{q}}}

\def\vs{{\bm{s}}}

\def\vv{{\bm{v}}}
\def\vw{{\bm{w}}}
\def\vx{{\bm{x}}}
\def\vy{{\bm{y}}}
\def\vz{{\bm{z}}}

\def\evc{{c}}

\def\evw{{w}}

\def\mB{{\bm{B}}}
\def\mC{{\bm{C}}}

\def\mI{{\bm{I}}}

\def\mR{{\bm{R}}}

\def\mW{{\bm{W}}}
\def\mX{{\bm{X}}}

\DeclareMathAlphabet{\mathsfit}{\encodingdefault}{\sfdefault}{m}{sl}
\SetMathAlphabet{\mathsfit}{bold}{\encodingdefault}{\sfdefault}{bx}{n}
\newcommand{\tens}[1]{\bm{\mathsfit{#1}}}

\def\tB{{\tens{B}}}

\def\emW{{W}}

\newcommand{\E}{\mathbb{E}}

\newcommand{\R}{\mathbb{R}}

\newcommand{\Var}{\mathrm{Var}}

\usepackage{mathtools}
\usepackage{xparse}
\usepackage{amssymb}
\usepackage{bm}
\usepackage{xfrac}
\usepackage{tensor}
\usepackage{eqparbox}
\definecolor{mathJustification}{rgb}{0.035, 0.298, 0.345}
\newcommand{\justification}[1]{{\color{mathJustification} #1}}

\DeclarePairedDelimiterX{\norm}[1]{\lVert}{\rVert}{#1}

\newcommand{\N}{\mathbb{N}} 
\newcommand{\Identity}[1][]{\mI_{#1}} 
\newcommand{\transpose}{\intercal}

\newcommand{\st}{\,|\,}

\newcommand{\g}[1][]{g_{#1}}
\newcommand{\ginv}[1][]{\g[#1]^{\text{-}1}}

\newcommand{\impg}[1][]{\overline{g}_{#1}}

\newcommand{\G}{\mathcal{G}} 
\newcommand{\impG}{\overline{\mathcal{G}}} 
\newcommand{\KleinFour}{\mathcal{K}_{4}} 
\newcommand{\Cyclic}[1][]{\mathcal{C}_{#1}} 
\newcommand{\Dihedral}[1][]{\mathcal{D}_{#1}} 

\newcommand{\EG}[1][\dimfiber]{\mathbb{E}_{#1}} 
\newcommand{\SEG}[1][n]{\mathbb{SE}_{#1}}

\newcommand{\rep}[2][]{
    {\rho_{\scriptscriptstyle{#1}}
    \def\temp{#2}\ifx\temp\empty
    \else
      (#2)%
    \fi
    }
}

\newcommand{\world}{\mathit{o}}
\newcommand{\refworld}{\overline{\mathit{o}}}
\newcommand{\pos}[1][]{\mathbf{r}_{#1}}           
\newcommand{\vel}[1][]{\mathbf{\dot{r}}_{#1}}     
\newcommand{\angvel}[1][]{\boldsymbol{\mathit{w}}_{#1}}  
\newcommand{\gvel}[1][]{\mathbf{\dot{r}}_{g,#1}} 
\newcommand{\gangvel}[1][]{\boldsymbol{\mathit{w}}_{g,#1}}  
      
\newcommand{\refvel}[1][]{\dot{\overline{\mathbf{r}}}_{#1}} 
\newcommand{\refangvel}[1][]{\overline{\boldsymbol{\mathit{w}}}_{#1}}  
 
\newcommand{\PosJacob}[1][]{\mathbf{J}_{T_{#1}}} 
\newcommand{\OriJacob}[1][]{\mathbf{J}_{R_{#1}}} 
\newcommand{\momentum}[1][]{\vh_{#1}}   
\newcommand{\angMomentum}[1][]{\rvk_{#1}}   
\newcommand{\linMomentum}[1][]{\vl_{#1}}   
\newcommand{\CMM}{\mathbf{A}_{G}}   
\newcommand{\bodyIdx}{i}
\newcommand{\impgBodyIdx}{k}
\newcommand{\refFrom}[2]{\tensor*[^{#1}]{{#2}}{}}
\newcommand{\refToFrom}[3]{\tensor*[#1]{{#2}}{#3}}     

\newcommand{\q}[1][]{\vq_{#1}}
\newcommand{\dq}[1][]{\dot{\vq}_{#1}}
\newcommand{\ddq}[1][]{\ddot{\vq}_{#1}}
\newcommand{\gq}[1][\g]{#1\cdot\q}
\newcommand{\gdq}[1][\g]{#1\cdot\dq}
\newcommand{\gddq}[1][\g]{#1\cdot\ddq}

\newcommand{\qj}{\hat{\vq}}           
\newcommand{\dqj}{\dot{\hat{\vq}}}    
\newcommand{\dimfiber}{d}             
\newcommand{\nq}{n}             
\newcommand{\nj}{{n_{J}}}             
\newcommand{\nb}{{n_{B}}}             

\newcommand{\Rotation}[1][]{{\mR_{#1}}}

\newcommand{\base}{B}
\newcommand{\smallBase}{{\base}}

\newcommand{\memSE}[1][]{\mX_{#1}}

\newcommand{\tanSE}[1][]{\dot{\mX}_{#1}}

\newcommand{\baseSE}{\memSE[{\smallBase}]}
\newcommand{\basese}{\tanSE[{\smallBase}]}

\newcommand{\ConfSpace}{\mathrm{Q}} 
\newcommand{\ConfSpaceJS}{\mathrm{Q}_{J}} 
\newcommand{\TangConfSpace}{\mathrm{T}_{\q}\mathrm{Q}} 
\newcommand{\Lagrangian}[1][\q, \dq]{
    \mathcal{L}
    \def\temp{#1}\ifx\temp\empty
    \else
      \left(#1\right)%
    \fi
    }

\newcommand{\KinE}[1][\q, \dq]{\mathcal{T}(#1)}   
\newcommand{\PotGE}[1][\q]{\mathcal{U}(#1)}   
\newcommand{\Mass}{\mathbf{M}}

\newcommand{\Inertia}[1][]{\mathbf{I}_{#1}}
\newcommand{\RefInertia}[1][]{\overline{\mathbf{I}}_{#1}}
\newcommand{\mass}[1][]{m_{#1}}
\newcommand{\refmass}[1][]{\overline{m}_{#1}}

\newcommand{\genForces}{\boldsymbol{\tau}}

\newcommand{\PrincipalFrame}[1][]{p_{#1}}

\newcommand{\nnParams}{\boldsymbol{\phi}}
\newcommand{\nn}[2][\nnParams]{\hat{f}\left(#2;#1\right)}
\newcommand{\nnIn}[1][]{\vx_{#1}}
\newcommand{\nnOut}[1][]{\vy_{#1}}
\newcommand{\nnInSpace}[1][]{\mathcal{X}}
\newcommand{\nnOutSpace}[1][]{\mathcal{Y}}
\newcommand{\nnInRand}[1][]{\rx_{#1}}
\newcommand{\nnOutRand}[1][]{\ry_{#1}}
\newcommand{\nnW}[1][]{\mW_{#1}}
\newcommand{\nnWEV}[1][]{\emW_{#1}}
\newcommand{\nnWflatDef}[1][]{vec(\mW_{#1})}
\newcommand{\nnWflat}[1][]{
    \ifthenelse{\isempty{#1}}
    {\vw_{#1}}
    {\evw_{#1}}
}
\newcommand{\nnWflatEV}[1][]{\evw_{#1}}
\newcommand{\nnBias}[1][]{\bm{b}_{#1}}
\newcommand{\nnAct}{\sigma}
\newcommand{\constraintMatrix}{\mC}
\newcommand{\basisEquiv}[1][]{
    {\mB_{#1}}
}
\newcommand{\basisEquivT}[1][]{\tB_{#1}}
\newcommand{\basisCoefV}[1][]{\vc_{#1}}
\newcommand{\basisCoefEV}[1][]{\evc_{#1}}

\newcommand{\basisRank}{r}
\newcommand{\sumBasisSquared}{\lambda_{\basisEquivT}}
\newcommand{\layerIdx}{l}
\newcommand{\invtrace}[2][]{\chi^{\boldsymbol{1}}_{#1}(#2)}

\usepackage{tcolorbox} 
\usepackage{xcolor}

\usepackage{physics}
\usepackage{amsmath}
\usepackage{tikz}
\usepackage{mathdots}
\usepackage{yhmath}
\usepackage{cancel}
\usepackage{color}
\usepackage{siunitx}
\usepackage{array}
\usepackage{multirow}
\usepackage{amssymb}
\usepackage{gensymb}
\usepackage{tabularx}
\usepackage{extarrows}
\usepackage{booktabs}
\usetikzlibrary{fadings}
\usetikzlibrary{patterns}
\usetikzlibrary{shadows.blur}
\usetikzlibrary{shapes}

\newtcolorbox{mybox}[3][]
{
  colframe = #2!25,
  colback  = #2!10,
  coltitle = #2!20!black,  
  title    = {#3},
  #1,
}

\definecolor{awesomeblue}{rgb}{0.054, 0.415, 0.505}
\definecolor{awesomeorange}{rgb}{0.570, 0.458, 0.0912}

\newtcolorbox{coloredFrame}[3][]
{
  colframe = #2!15,
  colback  = #2!15,
  coltitle = #2!20!black,  
  title    = {#3},
  #1,
}

\newcommand{\ubcolor}[3]{{
        \color{#1}{
            \underbrace{\color{black}{#2}}_{#3}
        }
    }}

\usepackage{pifont}
\usepackage{hyperref}
\usepackage{url}
\usepackage[nameinlink]{cleveref}  
\graphicspath{images}

\begin{document}

\title{
       On discrete symmetries of robotics systems:\\
       A group-theoretic and data-driven analysis
       }




%
\author{
\authorblockN{
    Daniel Ordonez-Apraez\authorrefmark{1}\authorrefmark{2},
    Mario Martin\authorrefmark{3}\authorrefmark{4},
    Antonio Agudo\authorrefmark{2} and
    Francesc Moreno-Noguer\authorrefmark{2}
}
\authorblockA{\footnotesize
    \authorrefmark{1} Istituto italiano di tecnologia IIT, Dynamic Legged Systems IIT-DLS \& Computational Statistics and Machine Learning IIT-CSML\\ \authorrefmark{2} Institut de Rob\`otica i Inform\`atica Industrial CSIC-UPC. \authorrefmark{3} Universitat Politècnica de Catalunya UPC. \authorrefmark{4} Barcelona Supercomputing Center BSC\\
}
\authorblockA{
    {\tt\footnotesize daniel.ordonez@iit.it, [aagudo, fmoreno]@iri.upc.edu,  mmartin@cs.upc.edu}
} 
}

\maketitle

%
\begin{abstract}
We present a comprehensive study on discrete morphological symmetries of dynamical systems, which are commonly observed in biological and artificial locomoting systems, such as legged, swimming, and flying animals/robots/virtual characters. These symmetries arise from the presence of one or more planes/axis of symmetry in the system's morphology, resulting in harmonious duplication and distribution of body parts. Significantly, we characterize how morphological symmetries extend to symmetries in the system's dynamics, optimal control policies, and in all proprioceptive and exteroceptive measurements related to the system's dynamics evolution. In the context of data-driven methods, symmetry represents an inductive bias that justifies the use of data augmentation or symmetric function approximators. To tackle this, we present a theoretical and practical framework for identifying the system's morphological symmetry group $\G$ and characterizing the symmetries in proprioceptive and exteroceptive data measurements. We then exploit these symmetries using data augmentation and $\G$-equivariant neural networks. Our experiments on both synthetic and real-world applications provide empirical evidence of the advantageous outcomes resulting from the exploitation of these symmetries, including improved sample efficiency, enhanced generalization, and reduction of trainable parameters.
\end{abstract}
\IEEEpeerreviewmaketitle

\section{Introduction}
    \begin{figure*}[t!]
        \centering
        \def\svgwidth{\textwidth}
        \input{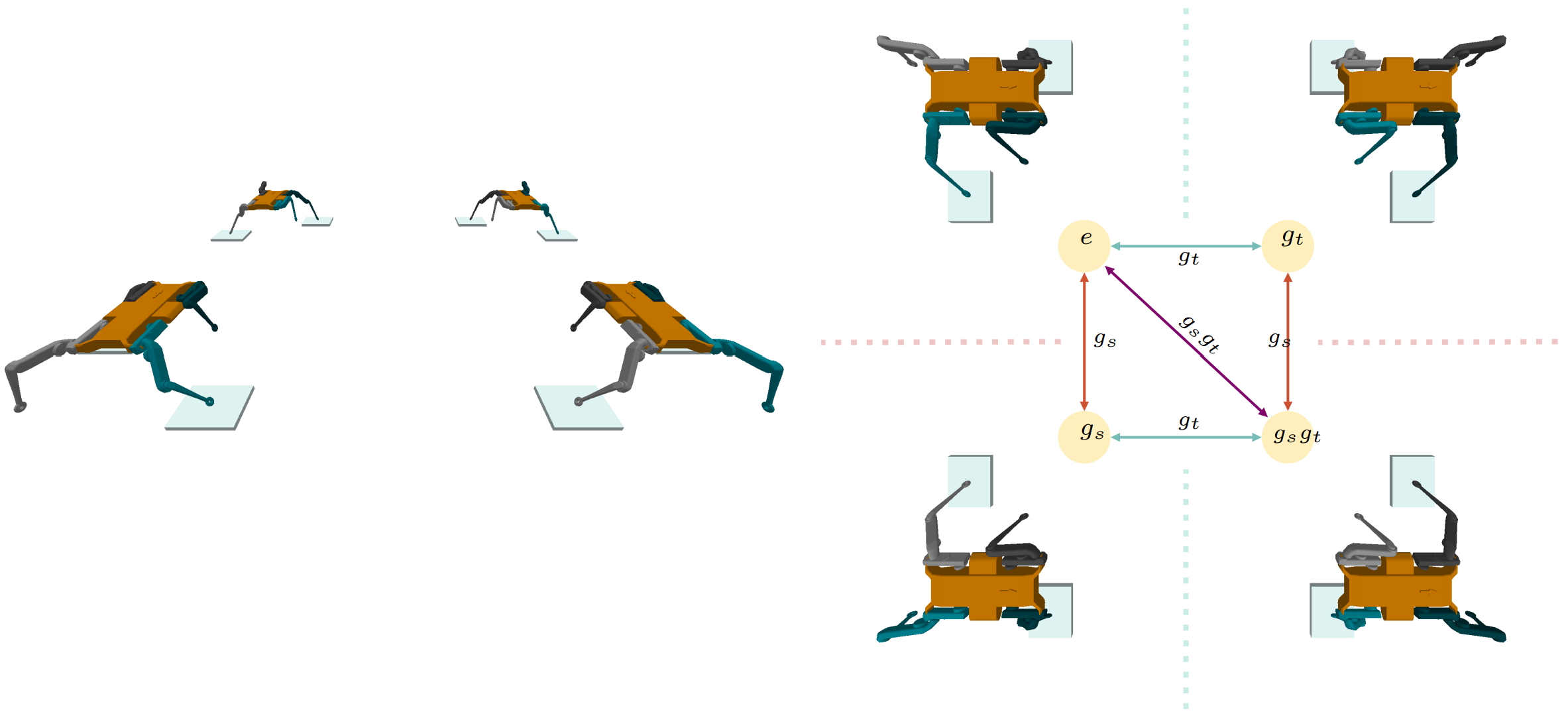}
        \vspace*{-5mm}
        \caption{
            \textbf{Left:} Symmetric configurations of the bipedal robot Atlas (\href{https://bit.ly/3HTn7bM}{3D animation}) illustrating its morphological symmetry described by the reflection group $\Cyclic[2]$. The robot can imitate reflections $\g[s]$ (hint: note the non-reflected text on the chest). \textbf{Middle:} Top-view of symmetric configurations of the quadruped robot Solo (\href{https://bit.ly/3wSzjDd}{3D animation}) showcasing its morphological symmetries described by the Klein four-group $\KleinFour$. The robot can imitate two perpendicular reflections $(\g[s], \g[t])$ and a $180^\circ$ rotation ($\g[r]$) of space (hint: observe the unreflected/unrotated robot's heading direction and legs coloring). Symmetry transformations (arrows) affect the robot's configuration, as well as proprioceptive measurements (center of mass linear $\linMomentum$ and angular $\angMomentum$ momentum) and exteroceptive measurements (terrain elevation, external force $\bm{f_1}$). \textbf{Right:} Diagram of a toy $\KleinFour$-equivariant neural network, processing the symmetric states of robot Solo $\nnIn$ and outputting the symmetric binary foot contact states $\nnOut$ (see \cref{sec:machine_learning_symmetries}).
        }
        \vspace*{-4mm}
        \label{fig:teaser}
    \end{figure*}
     Discrete Morphological Symmetries (DMSs) are ubiquitous in both biological and robotic systems. The vast majority of living and extinct animal species, including humans, exhibit bilateral/sagittal reflection symmetry, where the right side of the body is approximately a reflection of the left side (see \cref{fig:teaser}-left). Similarly, a significant number of species exhibit radial symmetry, characterized by two or more morphological symmetry planes/axis (see \cref{fig:teaser}-center) \citep{hollo2017demystification_symmetries_in_nature}. These symmetries are a consequence of nature's tendency to symmetric body parts and harmonic duplication and distribution of limbs. A  pattern perfected and exploited in the design of robotic systems. 
     
     To exploit morphological symmetries for control, learning, and computational design, it is necessary to establish a rigorous definition of morphological symmetry within the framework of dynamical systems theory. In \cref{sec:discrete_morphological_symmetries}, we define a DMS as an energy-preserving linear transformation of the system state configuration, which allows the system to \justification{imitate} some reflection, rotation, or translation of space. For instance, see how the bipedal robot Atlas and the quadruped Solo in \cref{fig:teaser} \justification{imitate} the reflection of space ($\g[s]$) with a discrete change in their body and limbs pose (state configuration). The existence of a DMS is subjected to constraints in the system's morphology, which manifest in identifiable symmetry constraints of the system's generalized mass matrix (\cref{sec:symmetries_of_dynamical_systems}).

    Symmetries of the state-space of a dynamical system translate to symmetries of the system's dynamics and control \citep{zinkevich2001symmetry_mdp_implications}. Thus, DMSs imply the presence of symmetries in the dynamics and control of body motions, resulting in symmetries in all proprioceptive and exteroceptive measurements related to the evolution of the system's dynamics (e.g., joint position/velocity/torque, depth images, contact forces). This property, in data-driven applications, opens the door for the use of data augmentation to mitigate challenges of data collection in the fields of robotics, computer graphics, and computational biology. Similarly, the use of symmetry constraints in machine learning algorithms is a known technique to enhance generalization and sample efficiency, while reducing the number of trainable parameters \citep{zinkevich2001symmetry_mdp_implications,finzi2021practical,van2020mdp}.

    Despite the potential benefits of exploiting symmetry and the ubiquitous presence of morphological symmetries in robotic/biological/virtual systems, this relevant inductive bias is frequently left unexploited in data-driven applications in robotics, computational biology, and computer graphics. We attribute the scarce adoption of these techniques to a missing theoretical framework that consolidates the concept of morphological symmetries, facilitating their study and identification. And, to a missing practical framework enabling the efficient and convenient exploitation of symmetries in real-world data-driven applications. 

    The identification of morphological symmetries and how these extend to symmetries of proprioceptive and exteroceptive data is currently a laborious and error-prone system-specific process, due to the lack of a clear theoretical framework. As a result, most recent works that exploit some morphological symmetry (e.g., \citep{yeh2019chirality, abdolhosseini2019symmetric_locomotion, yu2018learning} in computer graphics and \citep{van2020mdp, ordonez2022adaptable, hamed2013event_lef_right, finzi2021residualPP} in robotics/dynamical systems) have only been applied to simple systems and the simplest morphological symmetry: reflection/sagittal symmetry (see \cref{fig:teaser}-left), with the exception of \citet{finzi2021residualPP}. However, these works provide little guidance on how to apply these techniques to other systems, particularly those with more than a single morphological symmetry.
    
    Contrary to previous works, this paper focuses on understanding and exploiting morphological symmetries in arbitrary dynamical systems, with any number of symmetries.  To achieve this, we study morphological symmetries from the lens of dynamical systems and of group theory (the field of mathematics that studies symmetries, broadly used in machine learning and physics) (\cref{sec:background,sec:symmetries_of_dynamical_systems}). In summary, our work presents the following theoretical contributions:
    \begin{itemize}
        \item[\ding{104}] Identification of the set of DMSs of a dynamical system as a \justification{symmetry group} $\G$, that is \href{https://mathworld.wolfram.com/IsomorphicGroups.html}{isomorphic} to a group of isometries of the Euclidean space (\cref{sec:discrete_morphological_symmetries}).
        \item[\ding{104}] Characterization of DMSs as transformations to which the system's generalized mass matrix is equivariant. Enabling algorithmic identification of $\G$  (\cref{sec:symmetries_of_dynamical_systems}). 
        \item[\ding{104}] Characterization of how symmetries in proprioceptive and exteroceptive measurements arise from DMSs. 
    \end{itemize}
    Furthermore, our \justification{practical} contributions (\cref{sec:machine_learning_symmetries}) are: 
    \begin{itemize}
        \item[\ding{69}] An open-access repository\footnote{
            \label{foot:code} \href{https://bit.ly/44Dykqq}{github.com/Danfoa/MorphoSymm}
        } with example dynamical systems with DMSs, and the tools to prototype large-scale $\G$-equivariant Neural Networks (NN) for arbitrary DMSs. 
        \item[\ding{69}] Proof of an approximate $\sfrac{1}{|\G|}$ reduction in the trainable parameters of NN. Being $|\G|$ the number of DMSs.
        \item[\ding{69}] Derivation of an optimal initialization for the trainable parameters of $\G$-equivariant NN layers. 
    \end{itemize}
    Lastly, we provide optional appendices where we extend our theoretical derivations and provide tutorial-like examples.

\section{Background on Symmetry Groups}
    \label{sec:background}
    %
    Group theory is the default language for studying symmetry transformations. Thus, we provide a shallow introduction to the field\footnote{
        Being this short section undoubtedly an unsatisfactory introduction to group theory, we refer the uninitiated and interested reader to \citet{carter2021visual} for intuition and to \citet{bronstein2021geometric} for a machine learning introduction.
    } and define the notation required for our development.
    In a nutshell, a symmetry group in group theory is an abstraction of the \justification{set} of symmetries that \justification{different} geometric objects have. Understanding symmetry as a \justification{transformation} that conserves a relevant property of the object (e.g. energy). 
    
    For instance, in \cref{fig:teaser}-left the reflection group $\Cyclic[2]$ describes the symmetries that vectors, pseudo-vectors, rigid-bodies, and the robot Atlas have to a reflection of space $\g[s]$. Being the symmetries, transformations that preserve vector magnitudes and the robot's energy. Similarly in \cref{fig:teaser}-center, the Klein four-group $\KleinFour$
    describes the symmetries that the quadruped robot Solo has to $180^\circ$ rotations ($\g[r]$) and two perpendicular reflections $(\g[s], \g[t])$.  While on \cref{fig:teaser}-right the same group describes the symmetries of the input $\nnIn$ and output $\nnOut$ vector spaces of a $\KleinFour$-equivariant NN. 
    
    This formalism enables us to study the set of DMSs of a system and the set of symmetries of proprioceptive and exteroceptive data measurements as different representations of the \justification{same} symmetry group.  
    Formally, a symmetry group is a set of invertible symmetry transformations (or \justification{\textit{actions}}) $\G=\{e, \g[1], \ginv[1], \g[2], \dots\}$, including the trivial action $e$, which leaves objects unchanged. A group has an associative composition operator $(\cdot): \G \times \G \rightarrow \G$ mapping group actions to other group actions. Since two different geometric objects can share the same symmetry group, an action of the group must act differently on the two objects. Here, is where group \justification{\textit{representations}} allow us to use the familiar language of linear algebra to characterize how an action $g$ transforms a specific geometric object, say $\nnIn \in \nnInSpace \subseteq \R^{k}$. A representation $\rho_{\nnInSpace}: \G \rightarrow \mathcal{GL}(k)$ is a mapping from group actions to the set of invertible square matrices of $k$ dimensions (the General Linear group $\mathcal{GL}$). Thus, a representation specifies how objects $\nnIn \in \nnInSpace$ are transformed by group actions: $\g \cdot \nnIn \doteq \rep[\nnInSpace]{\g}\nnIn$. 
    
    A fundamental concept for this work is the notion of function $\G$-equivariance and $\G$-invariance. The function $f:\R^n \rightarrow \R^m$, is said to be $\G$-equivariant or $\G$-invariant if:
        \begin{equation}
            \ubcolor{awesomeblue}{
            \g \cdot \nnOut = f(\g \cdot \nnIn) \st \forall \g \in \G
            }{Equivariance}
            \quad \text{or} \quad
            \ubcolor{awesomeorange}{
            \nnOut = f(\g \cdot \nnIn) \st \forall \g \in \G
            }{Invariance}
            .
            \label{eq:equivariance-invariance-constraints}
        \end{equation}
    Roughly speaking, an equivariant function maps symmetries of the input to symmetries of the output, while an invariant function maps symmetries of the input to an invariant output. 

\section{Lagrangian Mechanics and Symmetries of Dynamical Systems}
    \label{sec:symmetries_of_dynamical_systems}
    Here we provide a group-theoretic perspective of symmetries in an arbitrary dynamical system. The definitions and notations of this section are fundamental for understanding the objective of this work, namely DMSs. To this end, let us consider a dynamical system with generalized coordinates $\q \in \ConfSpace \subseteq \R^{\nq}$ and velocities $\dq \in \TangConfSpace \subseteq \R^{\nq}$. Being $\ConfSpace$ the constrained configuration space, and $\TangConfSpace$ the space of constrained generalized velocities (i.e., the configuration tangent space at $\q$). Additionally, consider a Lagrangian function $\Lagrangian[]: \ConfSpace \times \TangConfSpace \rightarrow \R = \KinE[\q,\dq] - \PotGE[\q,\dq]$ specifying the energy state of the system at any state. Where $\KinE[\q,\dq]$, $\PotGE[\q,\dq]$ describe the state kinetic and potential energies, respectively. 

    The symmetries of a dynamical system are defined as transformations in the space of generalized coordinates that keep the energy of the system invariant \citep{ostrowski1996geometric_perspectives,lanczos2020variational_principles_mechanics}. In this work, we focus on time-invariant linear transformations
    of generalized coordinates $\rep[\ConfSpace]{\g}: \R^\nq \rightarrow \R^\nq$, which are  the representations of actions of a symmetry group: $\g \cdot \q \doteq \rep[\ConfSpace]{\g} \q \st \forall \; \q \in \ConfSpace, \g \in \G$. Note that because of the linearity of the transformation, the velocity and acceleration of the transformed coordinates are given by $\g \cdot \dq \doteq \rep[\ConfSpace]{\g} \dq $ and $\g \cdot \ddq \doteq \rep[\ConfSpace]{\g} \ddq$, respectively.
    
    Formally, we say that a dynamical system has a symmetry group $\G$ if its \justification{Lagrangian is $\G$-invariant}:
    \begin{equation}
        \begin{split}    
        \Lagrangian = \Lagrangian[\gq, \gdq] \st \forall g \in \G, \q \in \ConfSpace, \dq \in \TangConfSpace.\\
        \end{split}
        \label{eq:lagrangian-g_invariance}
    \end{equation}
    Being $\g$ a \justification{feasible} symmetry if the transformed state is a feasible state, i.e. when $\gq \ \in \ConfSpace$ and $\gdq \in \TangConfSpace$ (assuming both $\ConfSpace$ and $\TangConfSpace$ are connected sets).
    
    Since the Lagrangian structure differs between the original $(\q,\dq)$ and transformed coordinates $(\gq,\gdq) \;| \;\forall \;\g \in \G$, when we derive the Equations of Motion (EoM) of the system in the set of transformed coordinates, we obtain a set of EoMs describing  the system dynamics in different coordinate systems. Formally, if we derive the EoM through the Euler-Lagrange equation of the second order $\left(\frac{d}{dt} \frac{\partial \Lagrangian}{\partial \dq} - \frac{\partial \Lagrangian}{\partial \q} \equiv \Mass(\q)\ddq - \genForces(\q, \dq) = \bm{0}\right)$, the distinct EoM are equivariant
    to each other \citep{lanczos2020variational_principles_mechanics}, a property we will refer to as \justification{dynamics $\G$-equivariance}:
    \begin{multline}
        \g \cdot
        [
        \ubcolor{awesomeblue}{\Mass(\q)\ddq}{Inertial} - \ubcolor{awesomeblue}{\genForces(\q,\dq)}{Moving} 
        ]
        = 
        \ubcolor{awesomeorange}{\Mass(\gq)\gddq}{Inertial} - \ubcolor{awesomeorange}{\genForces(\gq,\gdq)}{Moving} = \bm{0} \\
        \st \; \forall \; \g \in \G,\;\q\in \ConfSpace, \;\dq\in \TangConfSpace.
        \label{eq:eom_g-equivariance}
    \end{multline}
    Denoting $\Mass(\q): \ConfSpace \rightarrow \R^{n \times n}$ as the generalized mass matrix function and $\genForces(\q,\dq): \ConfSpace \times \TangConfSpace \rightarrow \R^{n}$ as the generalized moving forces at a given state $(\q,\dq)$. Note that, in \cref{eq:eom_g-equivariance} the original and transformed dynamics are related linearly by the Jacobian of the coordinate transformation \citep{wheeler2014covariance_eom}. Which in this case is $\rep[\ConfSpace]{g}$ (reduced to $\g$ to preserve notation). 

    Note that to ensure dynamics $\G$-equivariance (\cref{eq:eom_g-equivariance}), both the generalized inertial and moving forces need to be independently equivariant, implying:
    \begin{multline}
        \Mass(\gq) = \g\Mass(\q)\ginv 
        \quad\land\quad
        \g \cdot \genForces(\q,\dq) = \genForces(\gq, \gdq) \\ 
        \st \; \forall \; \g \in \G,\; \q\in \ConfSpace, \; \dq\in \TangConfSpace.
        \label{eq:inertial_moving_forces_equivariance}
    \end{multline}
    The resultant $\G$-equivariance of the generalized mass matrix becomes an identifying property of symmetrical systems, providing a pathway for the \justification{identification} of action representations of the symmetry group $\rep[\ConfSpace]{\g} \;|\; \g \in \G$ (\cref{sec:dms_rigid_body_dynamics}). While the equivariance of the generalized moving forces (which in practice usually incorporates control, constraint, and external forces) implies that dynamics $\G$-equivariance (\cref{eq:eom_g-equivariance}) is upheld until a \justification{symmetry breaking force} violates the equivariance of $\genForces$. 

    To gain some intuition, consider as an example the bipedal robot Atlas, with symmetry group $\G=\Cyclic[2]=\{e,\g[s]\}$. According to \cref{eq:lagrangian-g_invariance} both robot states in \cref{fig:teaser}-left are symmetric states (related by the action $\g[s]$). Then, \cref{eq:eom_g-equivariance} suggests that any trajectory of motion, starting from the left robot state, will be equivalent (up to transformation by $\g[s]$) to a motion trajectory starting from the right robot state, if and only if, the moving forces driving both trajectories are equivalent (up to transformation by $\g[s]$). That is if the control and external forces are $\Cyclic[2]$-equivariant (\cref{eq:inertial_moving_forces_equivariance}). Note, we can perform a similar analysis for each symmetric state and action of systems with larger symmetry groups (e.g. Solo in \cref{fig:teaser}-center). 
    \subsubsection*{Floating-base dynamical systems}
        \label{sec:symmetries_floating_base}
        All robotic, biological, and virtual systems that move in a Euclidean space of $d$ dimensions, can be modeled as floating-base dynamical systems. Hence, without loss in generality, we assume   
        the system's configuration space can be decoupled into $\ConfSpace \doteq \EG[\dimfiber] \times \ConfSpaceJS$, being $\EG[\dimfiber]$ the space of all possible base configurations (all rotations/reflections and translations), and $\ConfSpaceJS$ the joint-space (or internal configuration space). Resulting in the decoupling
            $\q = \begin{bsmallmatrix}
                        \baseSE \\ \qj
                    \end{bsmallmatrix} 
                    \begin{smallmatrix}
                        \in &\EG[\dimfiber] \\ \in &\ConfSpaceJS
                    \end{smallmatrix}$
                    .
        Where $\baseSE$ is a homogenous matrix describing the base position and orientation in $\dimfiber$ dimensions
        \footnote{
            We use the homogeneous matrix representation of $\baseSE$ instead of a vector-quaternion representation, with some abuse of notation.
        }
        , and $\qj \subseteq \R^\nj$ represents the internal Degrees of Freedom (DoF) configuration. This separation of the configuration space becomes useful to study the effect of a symmetry transformation, since, we decouple the effect of the symmetry actions   
        $
            \gq = \rep[\ConfSpace]{\g} \: \q 
            = \begin{bsmallmatrix} 
                            \rep[{\EG[\dimfiber]}]{\g} & \boldsymbol{0}\\ 
                            \boldsymbol{0} & \rep[\ConfSpaceJS]{\g}
            \end{bsmallmatrix} 
            \begin{bsmallmatrix}
                \baseSE \\ \qj
            \end{bsmallmatrix}
            \; \big| \; \forall \; \g \in \G 
        $.
        With $\rep[{\EG[\dimfiber]}]{\g} \in \EG[\dimfiber]$ and $\rep[\ConfSpaceJS]{\g} \in \R^{\nj \times \nj}$ being representations of how action $\g$ transforms the base and joint-space configuration. 

    %
    \subsubsection*{Symmetries due to Euclidean isometries}
    \label{sec:continuous_symmetries_floating_base}
        floating-base systems are known for having symmetries to (some) translations, rotations, and reflections of space (i.e. \justification{\textit{Euclidean isometries}}). Giving origin to the conservation of linear/angular momentum, in conservative systems\citep{noether1918invariante}. We can understand these as:
        \begin{definition}[Symmetry due to Euclidean isometries]
        \label{def:symmetry_to_euclidean_isometry}
            A floating-base system with generalized coordinates $\q \in \ConfSpace \doteq \EG[\dimfiber] \times \ConfSpaceJS$, is said to be symmetric w.r.t a set of Euclidean isometries $\impg \in \impG \subseteq \EG[\dimfiber]$ (involving a true rotation, reflection or translation in space), if \cref{eq:lagrangian-g_invariance} holds for $\impG$. 
        \end{definition}
        Because rotations, reflections, and translations of space preserve the mass and inertia of bodies, as well as distances between them, these symmetries leave both the joint-space configuration and the generalized mass matrix \justification{invariant}: $\rep[\ConfSpaceJS]{\impg} = \Identity[\nj] \; \text{and} \;\Mass(\impg \cdot \qj) = \Mass(\qj) \st \; \forall \; \impg \in \impG \subseteq \EG[\dimfiber]$.
\section{Discrete morphological symmetries (DMSs)}
    \label{sec:discrete_morphological_symmetries}
    A dynamical system is said to possess a DMS if it can imitate the effects of a rotation, reflection, or translation in space through a feasible discrete change in its configuration. To gain some intuition, before introducing a formal definition, we can analyze the simplest and most common DMS.
    
    \subsubsection*{Reflection DMS}
    Although most floating-base dynamical systems are symmetric with respect to reflections of space (\cref{def:symmetry_to_euclidean_isometry}), these symmetries are infeasible due to the impossibility to execute reflections in the real-world \citep{selig2005geometric_fundamentals_robotics}. However, systems with sagittal symmetry (e.g., Atlas in \cref{fig:teaser}-left, or humans) can imitate the effect of a reflection with a feasible discrete change in their configuration, by rotating their body and modifying their limbs' pose. These systems share the same symmetry group, the reflection group $\G\equiv\Cyclic[2]$.
    \subsubsection*{Multiple DMSs} This property can be extended to the case of a floating-base system having multiple DMSs, allowing it to imitate multiple distinct Euclidean isometries. Most frequently systems can imitate a set of rotations and reflections, making $\G$ a Cyclic $\Cyclic[k]$ or Dihedral $\Dihedral[2k]$ group. See examples for $\Cyclic[3]$ in \cref{fig:examples_morphological_symmetries}, and for $\Dihedral[4] \equiv \KleinFour$ in \cref{fig:teaser}-center.
    
    We can formalize this property with as:
    \begin{definition}[Discrete morphological symmetry]
    \label{def:DMS}
        A floating-base dynamical system with generalized coordinates $\q \in \ConfSpace \doteq \EG[\dimfiber] \times \ConfSpaceJS$, is said to have a DMS if, for a given Euclidean isometry $\impg \in \EG[\dimfiber]$, there exists a \justification{feasible} action $\g \in \G$ 
        with a \justification{non-trivial} representation in joint-space ($\rep[\ConfSpaceJS]{\g} \neq \Identity[\nj]$), such that both $\g$ and $\impg$ are equivalent symmetries of the system:
        \begin{multline}
                \Lagrangian[\q,\dq] = 
                \Lagrangian[{\gq[\impg], \gdq[\impg]}] =  
                \Lagrangian[\gq,\gdq] 
                \\ \st \g \in \G,\; \impg \in \EG[\dimfiber], \; \forall \; \gq, \q \in \ConfSpace,\; \gdq, \dq \in \TangConfSpace.
            \label{eq:discrete_morphological_symmetries}
        \end{multline}
    \end{definition}
    The set of DMSs of the system forms its symmetry group $\G$. 
    Because each DMS is related with a system's symmetry $\impg$ due to a Euclidean isometry (\cref{def:symmetry_to_euclidean_isometry}), the group $\G$ is isomorphic to a subset of the Euclidean isometries of the system. 
    
    Hence, after identifying a potential Euclidean isometry to imitate $\impg \in \EG[\dimfiber]$, we can determine the DMS representation, considering that in any system state: 
    \begin{multline}
    \small
        \mathcal{L}\big(
            {\begin{bsmallmatrix} \rep[{\EG[\dimfiber]}]{\impg} \baseSE \\ \qj \end{bsmallmatrix}}, {\begin{bsmallmatrix} \rep[{\EG[\dimfiber]}]{\impg} \basese \\ \dqj \end{bsmallmatrix}}
        \big) 
        = 
        \mathcal{L}\big(
            {\begin{bsmallmatrix} 
                \rep[{\EG[\dimfiber]}]{\g} \baseSE\\ \rep[\ConfSpaceJS]{\g}\;\; \qj 
            \end{bsmallmatrix}}, 
            {\begin{bsmallmatrix} 
                \rep[{\EG[\dimfiber]}]{\g} \basese \\ \rep[\ConfSpaceJS]{\g}\;\; \dqj 
            \end{bsmallmatrix}}
        \big) 
        \\ 
        \bigg| \quad  
        \justification{ 
            \begin{aligned}
                &|\rep[{\EG[\dimfiber]}]{\impg}| = \pm 1, \; |\rep[{\EG[\dimfiber]}]{\g}| = 1
                \\
                &\rep[{\EG[\dimfiber]}]{\g} \memSE = \rep[{\EG[\dimfiber]}]{\impg} \memSE \rep[{\EG[\dimfiber]}]{\impg}^{\text{-}1}
            \end{aligned}        
        }
        \label{eq:diff_g_imp_g}
    \end{multline}
    Where the existence of the DMSs is subjected to the system's generalized mass matrix being $\G$-equivariant (\cref{eq:inertial_moving_forces_equivariance}), and to the transformation $\rep[{\EG[\dimfiber]}]{\g}$ (defined through group conjugation) being proper/feasible. In practice, these restrictions represent a pathway for the identification of $\G$ for any floating-base system (\cref{sec:identification_dms_g}). 
    %
    \subsection{Data augmentation in systems with DMS}
        \label{sec:dms_data_augmentation}        
        Recall from \cref{sec:symmetries_of_dynamical_systems} that due to the linearity, the action representation $\rep[\ConfSpace]{\g}$ acts on elements of configuration space $\ConfSpace$, configuration tangent space $\TangConfSpace$ and any higher order tangent spaces, including the spaces of generalized accelerations and forces  $\g \cdot \genForces = \rep[\ConfSpace]{\g}\genForces$ (\cref{eq:eom_g-equivariance}). Since for floating-base systems $\ConfSpace \doteq \EG[\dimfiber] \times \ConfSpaceJS$, this property translates to the action  representations on $\EG[\dimfiber]$ and $\ConfSpaceJS$. This effectively implies that $\rep[{\EG[\dimfiber]}]{\g}$ can be used to augment any point, vector, and orientation in $\EG[\dimfiber]$ and in $\EG[\dimfiber]$ higher order tangent spaces (e.g. locations of tactile sensing, linear \& angular velocities/accelerations, depth maps, external forces, terrain height-maps). Likewise the representation $\rep[{\ConfSpaceJS}]{\g}$ can be used to augment members of $\ConfSpaceJS$ and its higher order tangent spaces (e.g. joints positions/velocities/accelerations/torques). 

        In practice, this means that any proprioceptive and exteroceptive measurements relevant to the evolution of the system's dynamics can be augmented solely with combinations of $\rep[{\EG[\dimfiber]}]{\impg}$, $\rep[{\EG[\dimfiber]}]{\g}$ and $\rep[{\ConfSpaceJS}]{\g}$. Since these measurements consist of elements in $\ConfSpaceJS$, $\EG[\dimfiber]$, and their higher order tangent spaces (see examples in \cref{sec:sup_id_reps_com,sec:sup_id_reps_contact}). Furthermore, for any data point, there exist $|\G|$ symmetric data points (being $|\G|$ the order of the symmetry group). Therefore, for a system with a symmetry group of order, say, $|\G|=4$ (as in \cref{fig:teaser}-center), we can obtain an additional $3$ minutes of recordings for every minute of recorded data simply by considering the symmetric states of the data.
        
       To exploit the symmetries in the measurements, we first need to identify the joint-space representations $\rep[{\ConfSpaceJS}]{\g}$, which requires additional assumptions about the system's dynamics. In this work, we focus on the case of rigid-body dynamics, although a similar analysis can be extended to other types of systems, such as soft robots.
        %
        %
    %
    \subsection{DMS in the case of rigid-body dynamics} 
    \label{sec:dms_rigid_body_dynamics}
        Consider dynamical systems composed of $\nb$ interconnected rigid bodies evolving in $\EG$. This is the usual scenario in robotics, computer graphics, and computational biology. 
        
        The kinetic energy for these systems is determined by
        $
            \small
            \KinE = \frac{1}{2}\sum_{k}^{\nb} \mass[k]\vel[k]^2  + \angvel[k]^\transpose \Inertia[k] \angvel[k] = \frac{1}{2} \dq^{\transpose}\Mass(\q)\dq
        $. Being $\mass[k]$, $\Inertia[k]$, $\vel[k]$ and $\angvel[k]$ the mass, inertia, linear velocity, and angular velocity of body $k$. Considering that the energy-preservation property of symmetries ($\cref{eq:lagrangian-g_invariance}$) is dependent solely on the $\G$-equivariance of $\Mass(\q)$ (\cref{eq:inertial_moving_forces_equivariance}), we can assert the existence of DMSs by analyzing $\Mass(\q)$. The generalized mass matrix is given by $\Mass(\q) = \sum_{k}^{\nb} \PosJacob[k](\q)^\transpose \mass[k] \PosJacob[k](\q) + \OriJacob[k](\q)^\transpose \Inertia[k]\OriJacob[k](\q)$, being $\PosJacob[k](\q): \ConfSpace \rightarrow \R^{\dimfiber \cross \nq}$ and $\OriJacob[k](\q): \ConfSpace \rightarrow \R^{\dimfiber \cross \nq}$ the position and orientation Jacobians, used to map generalized velocities to the linear ($\vel[k] = \PosJacob[k](\q)\dq$) and angular ($\angvel[k] = \OriJacob[k](\q)\dq$) velocities of the body $k$ \citep{wieber2006holonomy}.
        
        These Jacobians are functions of the \justification{kinematic parameters} of the system\footnote{
            The Denavit–Hartenberg parameters are a common convention of kinematic parameters adopted in robotics and computer graphics.
        }. While the mass and inertia of all bodies are the system's \justification{dynamic parameters}. A DMS implies symmetries over both kinematic and dynamic parameters. 
        
        \subsubsection*{Symmetries of kinematic parameters (Kinematic Tree)} 
        The symmetry in kinematic parameters can be thought of as a kinematic tree symmetry. I.e., the DMS $\g$ must transform the system state in a way that produces a kinematic tree indistinguishable from the one obtained by applying the Euclidean isometry $\impg$. Ignoring the dynamic parameters makes it easier to see this.
        
        Consider that applying the Euclidean isometry $\impg$ conserves kinetic energy as velocity vectors are only rotated or reflected. Hence, for $\g$ to imitate the effect of $\impg$, the velocity of the $\impgBodyIdx^{th}$ body after applying $\impg$ must be equal to the velocity of the $\bodyIdx^{th}$ body after applying $\g$. In other words, 
        $
            \impg \cdot \vel[\impgBodyIdx] = \g \cdot \vel[\bodyIdx] \doteq \PosJacob[\bodyIdx](\g \cdot \q) \g \cdot \dq$ and 
            $ \impg \cdot \angvel[\impgBodyIdx] = \g \cdot \angvel[\bodyIdx] \doteq \OriJacob[\bodyIdx](\g \cdot \q) \g \cdot \dq \st \forall \; \{ (\impgBodyIdx, \bodyIdx) |\impgBodyIdx, \bodyIdx \in [\nb]\}
        $. 
        This results in the following position Jacobian constraints:
        \begin{align}
            \small
            \PosJacob[\bodyIdx](\gq) \g &= \impg \cdot \PosJacob[\impgBodyIdx](q) 
            \; \st \forall \q \in \ConfSpace, \{ (\impgBodyIdx, \bodyIdx) |\impgBodyIdx, \bodyIdx \in [\nb]\}
            \nonumber
            \\
            \PosJacob[\bodyIdx](\rep[\ConfSpace]{\g}\q) \rep[\ConfSpace]{\g} &= \rep[\EG]{\impg} \cdot \PosJacob[\impgBodyIdx](q). 
            \label{eq:kinematic_symmetries_rigid_bodies}
        \end{align}
         Analog constraints apply to the rotational Jacobian. 
         
         Equation \ref{eq:kinematic_symmetries_rigid_bodies} specifies the kinematic parameter constraints required for $\g$ to be a DMS, ensuring the $\g$-equivariance of $\Mass(\q)$.  Note that If $\impgBodyIdx \neq \bodyIdx$, the representation $\rep[\ConfSpaceJS]{\g}$ entails a permutation of the $\bodyIdx$ and $\impgBodyIdx$ joint-space configurations. For example, $\g[s]$ swaps the left and right leg configurations of Atlas and Solo in Figure \ref{fig:teaser}.
        \subsubsection*{Symmetries of dynamic parameters (Mass \& Inertia)}
        In addition to the kinematic tree symmetry of \cref{eq:kinematic_symmetries_rigid_bodies}, the mass, CoM, and inertia of the bodies $\bodyIdx$ and $\impgBodyIdx$ must be equivalent for $\g$ to preserve kinetic energy. To understand this morphological constraint, consider how, in \cref{eq:diff_g_imp_g}, the base body configuration $\baseSE \in \SEG[\dimfiber]$ is transformed by the DMS $\rep[{\EG[\dimfiber]}]{\g} \in \SEG[\dimfiber]$, and by the Euclidean isometry $\rep[{\EG[\dimfiber]}]{\impg} \in \EG[\dimfiber]$. For both base configurations to have the same dynamics, their CoM must coincide, and crucially, the reflected Inertia matrices $\Inertia[\base]$ need to be identical. This constraint is satisfied if the reflected Inertia is invariant to the transformation $\baseSE = \baseSE \rep[{\EG[\dimfiber]}]{\impg}^{\text{-}1}$. In practice, this invariance implies a \justification{symmetric mass distribution} of the rigid body (see geometric proof in \cref{sec:sup_reflection_symmetry}).

\begin{figure}[t!]
            \centering
            \includegraphics[width=\linewidth]{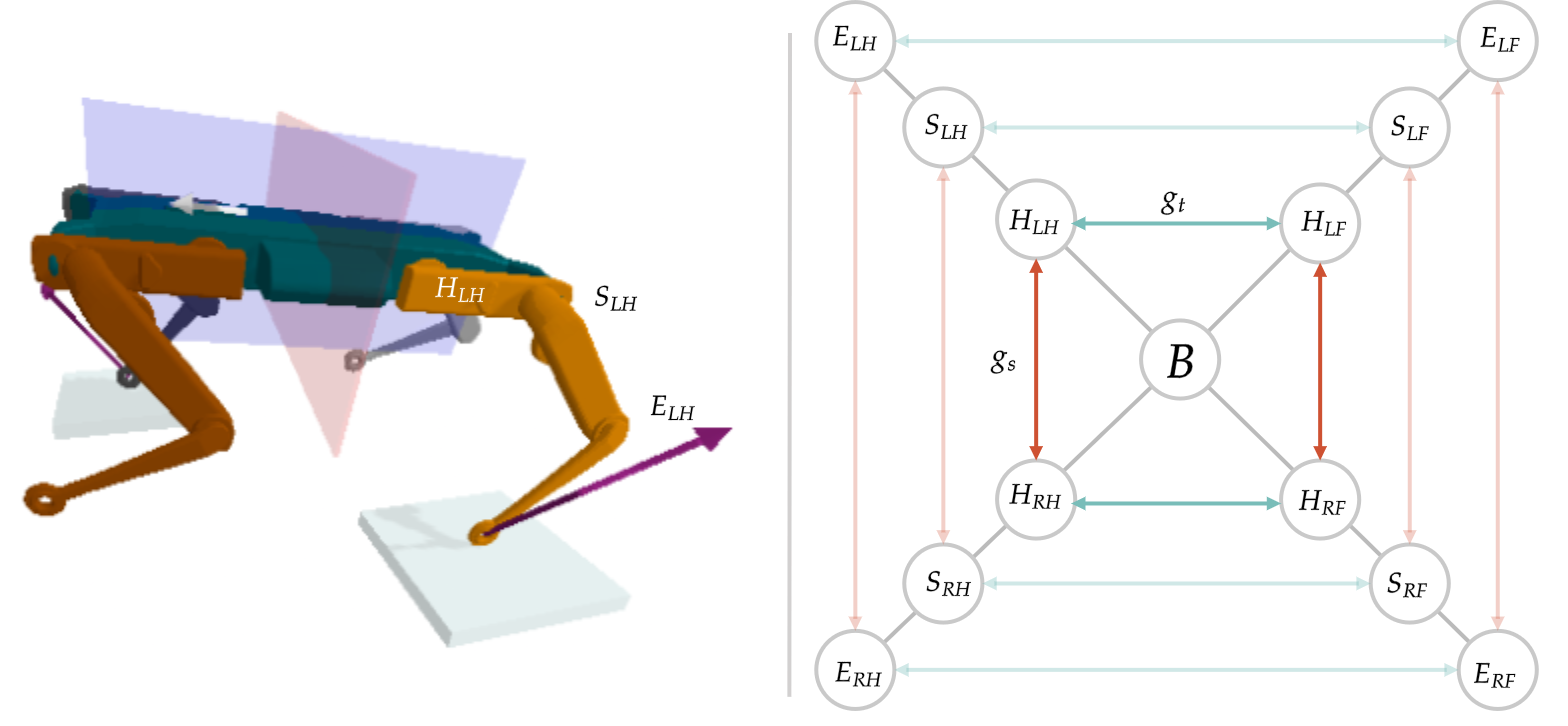}
            \caption{\textbf{Left:} Solo sagittal (blue) and transversal (red) symmetry planes of the base body. \textbf{Right:} Solo's kinematic tree, and permutation symmetries of the legs/tree-branches.}
            \vspace{-0.5cm}
            \label{fig:sup_solo_symmetry_planes}
        \end{figure}

         Let's consider the robot Solo in \cref{fig:teaser}-center as an example. It can imitate two reflections of space ($\impg[t]$, $\impg[s]$) and a $180^\circ$ rotation $\impg[r]$. The existence of the robot's DMSs can be attributed to two factors. First, the base body of Solo possesses two symmetry planes (\cref{fig:sup_solo_symmetry_planes}), leading to symmetric mass distributions and the invariance of the body's reflected inertia under the transformation $\baseSE \rep[\EG]{\impg}^{\text{-}1} \st \impg \in \KleinFour$. Second, the modularity of the kinematic tree. This arises from all four legs (tree branches) being structurally identical, resulting in the tree branches having equivalent inertial and kinematic parameters (see \cref{fig:sup_solo_symmetry_planes}). Consequently, we can define each $\rep[\ConfSpaceJS]{\g}$ as a permutation of the leg configurations. For instance, for $\g[s]$, the configurations of the left and right legs are interchanged, while for  $\g[t]$ the front and back legs' configuration is interchanged. It is important to note that this interchange of leg configurations would violate \cref{eq:lagrangian-g_invariance} if the legs were not composed of identical/reflected rigid bodies (i.e, having different mass distributions) and would violate \cref{eq:eom_g-equivariance} if the dynamics and constraints of the leg joints differed (e.g., different position/velocity/torque limits).
        
        %
        \subsection{Identification of DMS group $\G$ in rigid-body dynamics}
        \label{sec:identification_dms_g}
        The identification of the DMS group $\G$ in a floating-base dynamical system composed of rigid bodies can be achieved through the following four steps (see tutorial examples in \cref{fig:examples_morphological_symmetries}):
        \begin{enumerate}
            \item Identify all the unique bodies in the kinematic tree, including the base of the system, ensuring there are no duplicated or reflected versions of the same body.
            \item Determine the set of Euclidean isometries $\impg \in \EG[\dimfiber]$ for which the inertia of each unique body remains invariant. These candidate Euclidean isometries represent potential DMSs for the system.
            \item Identify any modularity in the kinematic tree, such as sets of duplicated or reflected symmetric kinematic subchains.
            \item Utilize \cref{eq:kinematic_symmetries_rigid_bodies} to assess the feasibility and existence of $\rep[\ConfSpaceJS]{\g} \in \G$, progressing from the base to the end-effectors.
        \end{enumerate}
        While the presented analysis may seem extensive for simple systems and DMS groups, these abstractions open avenues for studying more complex systems and symmetry groups, enabling computational design of symmetric robotic systems, as well as algorithmic identification of DMSs and data augmentation.
\section{$\G$-Equivariant function approximators}
    \label{sec:machine_learning_symmetries} 
    After identifying the DMS group $\G$ of our system (\cref{sec:identification_dms_g}) and understanding how these symmetries manifest in proprioceptive and exteroceptive measurements (\cref{sec:dms_data_augmentation}), we can now exploit these symmetries in our data.  Consider the symmetric input $\nnInSpace$ and output $\nnOutSpace$ vector spaces, of any $\G$-equivariant/invariant (\cref{eq:equivariance-invariance-constraints}) function $f: \nnInSpace \rightarrow \nnOutSpace$, that we desire to approximate with a model $\hat{f}$. 
    We assume that $\nnInSpace$ and $\nnOutSpace$ are symmetric spaces as they are composed of (potentially several) proprioceptive/exteroceptive measurements. For instance, both spaces could contain measurements of the system state, terrain elevation, external forces, depth images, contact states, etc. See \cref{sec:experiments} for examples. 
    
    To enhance the generalization and sample efficiency of our approximation, we can enforce the $\G$-equivariant/invariant constraints of the original function on our model $\hat{f}$. In this section, we outline the process of incorporating these constraints when $\hat{f}$ is a neural network parameterized by $\nnParams$. By imposing symmetry constraints, we can reduce the number of trainable parameters in the architecture. Our approach builds upon the theory and implementation of \citet{finzi2021practical}'s framework for $\G$-equivariant NNs. Our main motivation is to overcome the limitations that hinder the construction of large-scale $\G$-equivariant NNs, which are commonly encountered in real-life applications (see details in Section \ref{sec:sup_ml_contributions}).
    
    Consider c to be composed of multiple perceptrons (or convolutional) layers of the form $^{l}\nnOut: = \nnAct(^{l}\nnW  ^{l}\nnIn + \; ^{l}\nnBias)$, where $^{l}\nnIn\in \R^n$, $^{l}\nnOut \in \R^m$ are the $l^{th}$ layer input-output. $^{l}\nnW \in \R^{m \times n}$ and $^{l}\nnBias$ are the layer's linear map and bias; and $\nnAct: \R \rightarrow \R$ is a strictly monotonic nonlinearity \citep{ravanbakhsh2017equivariance_parameter_sharing}. With this parametrization, the equivariance constraints of \cref{eq:equivariance-invariance-constraints} can be reduced to constraints on the linear map $\nnW$  (dropping the layer index ${l}$ for notation clarity):\footnote{A similar analysis can be made for the bias vector $\nnBias$.}
    \begin{multline}
        \rep[out]{\g}\nnW = \nnW \rep[in]{\g}
        \;\; \iff \;\; 
        (\rep[\nnW]{\g} - \Identity) \nnWflat = \boldsymbol{0} 
        \\
        \st \forall \; \g \in \G
        .
        \label{eq:fix-point}
    \end{multline}
    The right-side of \cref{eq:fix-point} is a reformulation of the linear map equivariance constraints (left-side) as a standard set of linear equations. Denoting $\nnWflat = \nnWflatDef \in \R^{mn}$ as a vectorized version of $\nnW$ and $\rep[\nnW]{\g} = \rep[out]{\g} \otimes \rep[in]{\g^{\text{-}1}}^{\transpose} \in \R^{mn \times mn}$  as the action representation acting on the parameter space of the linear map ($\otimes$ stands for the Kronecker product). Here, we consider the group acting on $\nnW$ a \justification{semi-direct product}\footnote{\label{foot:semi-direct-product}
        Since with DMSs the input and output symmetry groups are isomorphic, using a direct product in \cref{eq:fix-point} implies an over-constraining of the linear map. Resulting in an excessive reduction in the number of trainable parameters.}
    of the input and output groups (refer to \citet{finzi2021practical} for details). 
    Since the constraint imposed by each $\g$ is linear in $\nnW$, we can stack them into a single large system of linear equations $\constraintMatrix \nnWflat = \boldsymbol{0}$. The nullspace of this system of equations $\basisEquiv \in \R^{mn \times \basisRank}$ describes the $\basisRank$ basis vectors spawning the entire space of equivariant linear maps. Allowing to parameterize all $\G$-equivariant $\nnW$ as: 
    \begin{equation}
        \begin{split}
            \nnWflat = {\textstyle \sum_{k}^{\basisRank}} \basisCoefEV[k] \basisEquiv[:,k] 
            \;\; \iff \;\; 
            \nnW &= 
                {\textstyle \sum_{k}^{r}} \basisCoefEV[k] \; \textrm{unvec}(\basisEquiv[:,k]) 
                \\
                &\doteq 
                {\textstyle \sum_{k}^{r}} \basisCoefEV[k]\basisEquivT[:,:,k].
        \end{split}
        \label{eq:equivariant_linear_maps_basis}
    \end{equation}
    Where the basis coefficients $\basisCoefV \in \R^\basisRank$ represent the free variables of the system of equations and the \justification{trainable parameters of the equivariant layer} (see \cref{fig:teaser} right).

\subsection{Dealing with memory complexity of equivariant layers:} 
    An equivariant layer needs to store the matrices $\rep[\nnW]{\g} \in \R^{mn \times mn}$ and $\basisEquiv \in \R^{mn \times r}$, in addition to the typical memory complexity of a perceptron or convolutional layer. These matrices' memory complexity quickly becomes intractable for moderate input-output dimensions (see \cref{table:cnn-ecnn_comparison}). Fortunately, the symmetry groups of DMSs (finite groups) have sparse action matrix representations, resulting in both of the aforementioned matrices being highly sparse. We extend the API from \citet{finzi2021practical} to handle sparse matrix representations, limiting the additional memory footprint to a minimum.

\subsection{Dealing with the computational complexity of determining the equivariant basis $\basisEquiv$} 
    Computing $\basisEquiv$ amounts to finding the nullspace of a large linear system of equations. \citet{finzi2021practical} proposes a Krylov gradient-based method able to handle both finite and Lie groups' arbitrary representations. While \citet{van2020mdp} approximates $\basisEquiv$ through SVD of a matrix $\bar\mW \in \R^{z \times mn}$ ($z \geq mn$). Both approaches run in polynomial time $\mathcal{O}(r^2(mn)^2)$ and approximate the space rank $\basisRank$ numerically. These approaches become intractable for large vector spaces. 
    
    To handle the computational cost we exploit the fact that for DMSs all $\rep[\ConfSpaceJS]{}$ and $\rep[{\nnOutSpace_l}]{}$ (representations of the output space of internal layers of the NN), can be expressed as permutation representations. Reducing the computation of the nullspace $\basisEquiv$ to a search of the permutations (or orbits) of each dimension of $\nnWflat$. A problem that can be solved in linear time. 
    
    For permutation representations, the constraints imposed by each $\rep[\nnW]{\g}$ on $\nnWflat$ can be interpreted as parameter sharing constraints. Becoming every unique orbit of the dimensions of the linear map $\G \cdot \nnWflat[k] = \{\g \cdot \nnWflat[k] \; :\; \forall \;\g \in \G \} = \{\nnWflat[k], -\nnWflat[i], \dots, \nnWflat[j]\}$, a vector of the null-space of $\constraintMatrix$ (i.e., $\basisEquiv[i]$). Each unique orbit describes the sharing scheme of a free variable of the system of equations, that is, the sharing of the trainable parameter $\basisCoefEV[i]$ over multiple positions in $\nnW$ (see the parameter orbits of length 4 in \cref{fig:teaser}-right, for $\KleinFour$). The orbits of all $w \in \nnWflat$ are trivially computed with $[\nnWflat, \rep[\nnW]{\g[1]}\nnWflat, \dots, \rep[\nnW]{\g[|\G|]}\nnWflat]$, while the unique $\basisRank$ orbits can be identified in $\mathcal{O}(mn)$ time. Our proposed solution can be thought of as a linear-time version of \citet{ravanbakhsh2017equivariance_parameter_sharing}.
\begin{figure*}[t!] 
    \centering
    \begin{subfigure}[b]{0.32\textwidth}
        \centering
       \includegraphics[width=\textwidth]{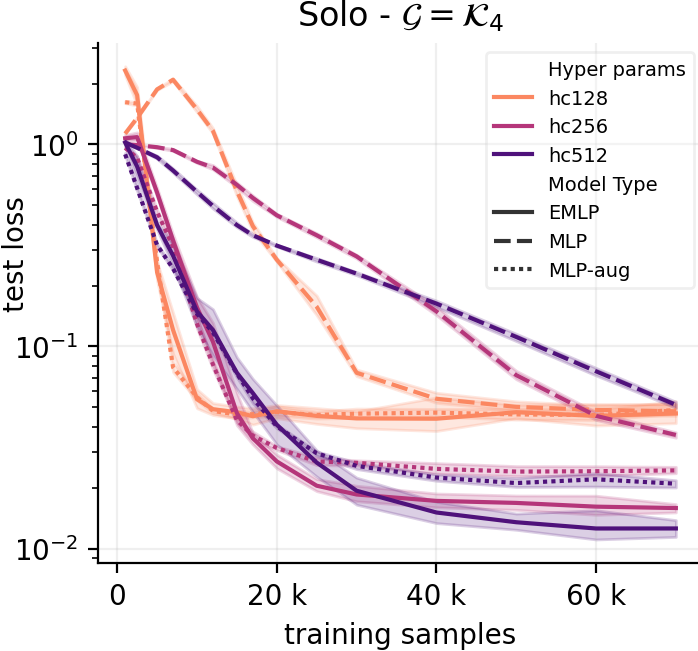}
       \label{fig:solo_results_k4_diff_sizes}
    \end{subfigure}
    \begin{subfigure}[b]{0.32\textwidth}
        \centering
       \includegraphics[width=\textwidth]{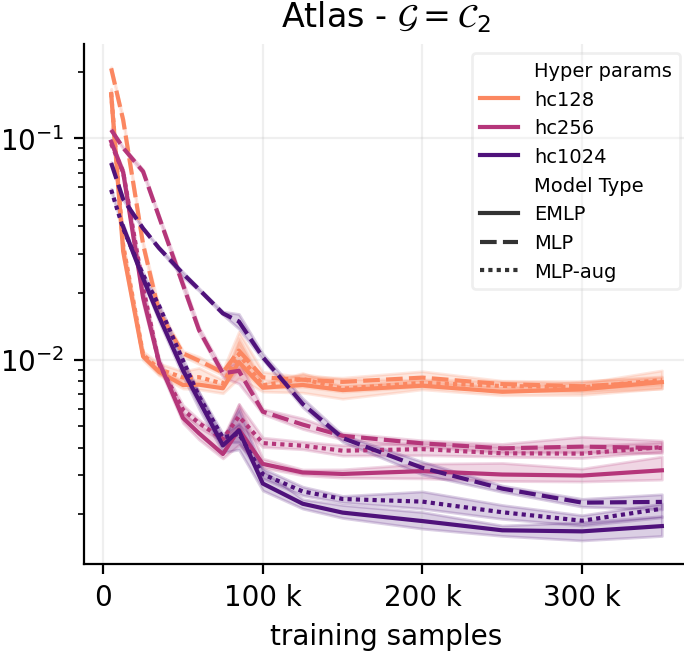}
       \label{fig:atlas_results_c2_diff_sizes}
    \end{subfigure}
    \begin{subfigure}[b]{0.32\textwidth}
       \centering
       \includegraphics[width=\textwidth]{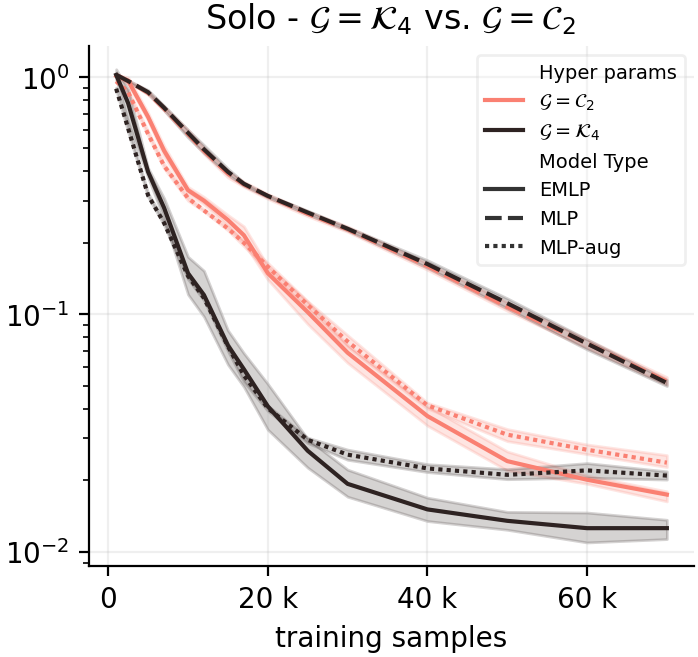}
       \label{fig:solo_results_k4_c2_comparison_512}
   \end{subfigure}
    \caption{
    \textbf{CoM-estimation results comparing MLP, MLP-aug, and EMLP models}. \textbf{Left and Middle:} Test set sample efficiency of model variants with different capacities (number of neurons \textit{hc} in hidden layers) for robot Solo and Atlas. \textbf{Right:} Sample efficiency for robot Solo with models having $hc=512$, when exploiting $\G=\KleinFour$ (sagittal and traversal symmetries) and $\G=\Cyclic[2] ={e, \g[s]} \subset \KleinFour$ (only sagittal symmetry). The plots depict the average and standard deviation across 10 seeds.
    \vspace{-0.50cm}
    }
    \label{fig:com_estimation_results}
\end{figure*}
\subsection{Optimal parameter initialization for equivariant layers}
\label{sec:param_init}%
        Proper initialization of the equivariant layer's trainable parameters $\basisCoefV[\layerIdx]$ (\cref{eq:equivariant_linear_maps_basis}) is required to avoid activations/gradients from vanishing or exploding \citep{klambauer2017self_normalizing_nn}. Following the same derivation of the Kaiming initialization \citep{he2015keiming_delving} (see \cref{sec:sup_initialization_equivariant_layers}), 
        we can conclude that the parameters should be initially sampled from a distribution with $\Var(\basisCoefV[\layerIdx]) = \sfrac{m}{\sumBasisSquared \gamma_{\nnAct}}$, to ensure constant variance of activations throughout the network layers (see \cref{fig:sup_initialization}). Where $\sumBasisSquared = \sum_{i}^{m}\sum_{j}^{n}\sum_{k}^{r} \basisEquivT[i:j:k]^2$ and $\gamma_{\sigma}$ is a nonlinearity dependant scalar (e.g., $\gamma_{\text{ReLu}}=\sfrac{1}{2}$, $\gamma_{\text{SeLu}}=1$ following \citet{klambauer2017self_normalizing_nn}). This initialization depends only on $\basisEquivT$. Thus, is applicable for any symmetry group.
\subsection{Reduction of trainable parameters in equivariant layers:} 
     Determining analytically the number of trainable parameters (i.e. the rank $\basisRank$) of an $\G$-equivariant layer is, in general, an unresolved problem. However, for DMS groups and permutation representations, it becomes trivial to show that the number of trainable parameters of a $\G$-equivariant layer can range from $\sfrac{|\nnWflat|}{|\G|} \leq r \leq |\nnWflat|$, depending on the number of dimensions of the input-output spaces left invariant by the symmetry actions (see details in \cref{sec:sup_parameter_reduction}). In practice, this implies that for a $\G$-equivariant layer without any input-output fixed points (e.g., all intermediate layers of a $\G$-equivariant NN), the number of trainable parameters is reduced by $\sfrac{1}{|\G|}$ being $|\G|$ the group order. Therefore a $\G$-equivariant architecture with $\G=\Cyclic[2]$ (or $\G=\KleinFour$) will have approximately \justification{$\sfrac{1}{2}$ (or $\sfrac{1}{4}$)} of the trainable parameters of an unconstrained NN of the same architectural size (this applies to NN processing data from robot Atlas and Solo \cref{fig:teaser}). The reduction of parameters is caused by the parameter sharing constraints (\cref{eq:equivariant_linear_maps_basis}) and is visually depicted in \cref{fig:teaser}-right.
     %
     %
\section{Experiments}
    \label{sec:experiments}
    We demonstrate the effectiveness of DMSs for data augmentation and training equivariant functions through two supervised learning experiments: a regression task using synthetic data and a classification task using real-world data. These experiments showcase the impact of exploiting DMSs on sample efficiency and generalization capacity. While we keep the presentation concise, all the technical aspects are detailed in \cref{sec:sup_experiment_details} and the code repository \footref{foot:code}.
        \subsection{CoM momentum estimation (Regression)} 
        In this experiment, we train a NN to approximate a robot's center-of-mass momentum given the joint-space position and velocities: $\momentum = \CMM(\qj)\dqj$, where $\momentum = [\linMomentum^{\transpose} \; \angMomentum^{\transpose}]^{\transpose}$ are the linear $\linMomentum$ and angular $\angMomentum$ momentum components
        and $\CMM$ is the Centroidal Momentum Matrix (CMM) of \citet{orin2013centroidal_momentum_matrix_cmm}. This analytical function is highly non-linear and $\G$-equivariant w.r.t the robot's symmetry group $\G$ (\cref{eq:com_momentum_pin,eq:com_momentum_pin_symm}). 
        
        We test two robots: Atlas, a $32$-DoF humanoid robot with $\G = \Cyclic[2]$ sagittal reflection symmetry (\cref{fig:teaser}-left), and Solo, a $12$-DoF quadruped robot with $\G=\KleinFour$ (\cref{fig:teaser}-center). We compare tree function approximation variants: a standard Multi-Layer Perceptron (MLP), an augmented MLP s(MLP-aug), and a hard-equivariant MLP (E-MLP).
    
        In \cref{fig:com_estimation_results}-left-\&-middle, we compare the model variants. Across both robots and all model capacities, E-MLP and MLP-Aug outperform MLP in terms of sample efficiency (better generalization with fewer data) and robustness to overfitting when training data is limited. Among the E-MLP and MLP-Aug variants, lower capacity versions exhibit similar behavior, but as capacity increases, E-MLP demonstrates superior sample efficiency and generalization. In addition, \cref{fig:com_estimation_results}-right shows a comparison for the Solo robot, evaluating the performance of the model variants when exploiting either the entire symmetry group ($\KleinFour$) or a subgroup of the true symmetry group ($\Cyclic[2] \subset \KleinFour$). The results indicate that sample efficiency and generalization capacity increase with the number of \justification{true} symmetries of the data exploited. 

        \subsection{Static-friction-regime contact detection (Classification)}
        \begin{figure*}[t!] 
            \centering
            \begin{subfigure}[b]{0.325\textwidth}
                \centering
                \includegraphics[width=\textwidth]{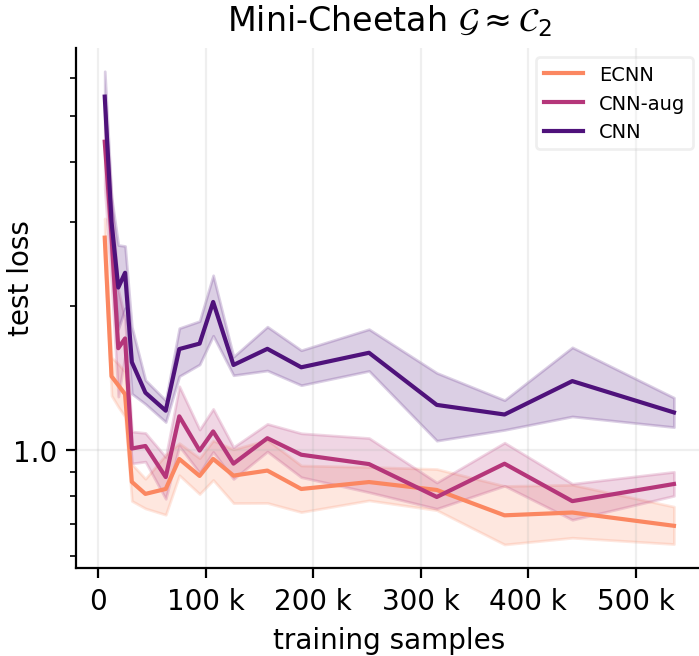}
            \end{subfigure}
            \begin{subfigure}[b]{0.325\textwidth}
               \centering
               \includegraphics[width=\textwidth]{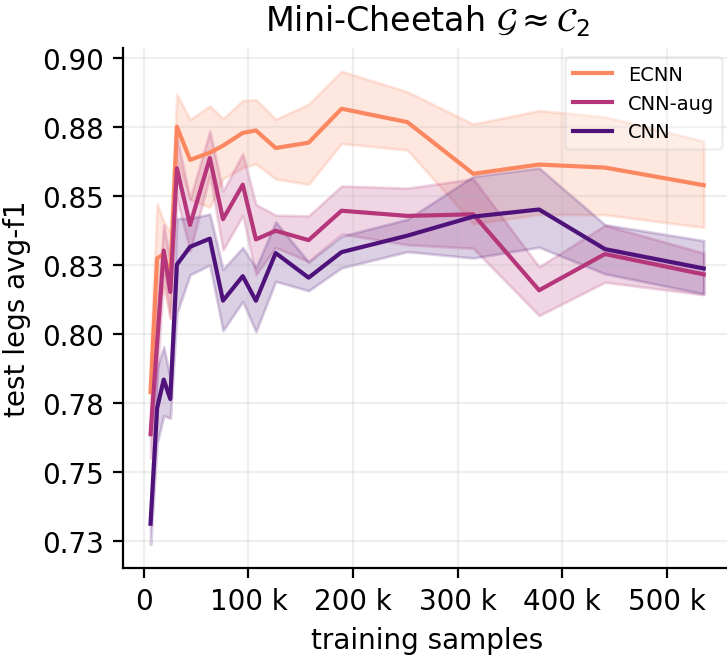}
            \end{subfigure}
            \hfill
            \begin{subfigure}[b]{0.325\textwidth}
                \centering
                \includegraphics[width=\textwidth] {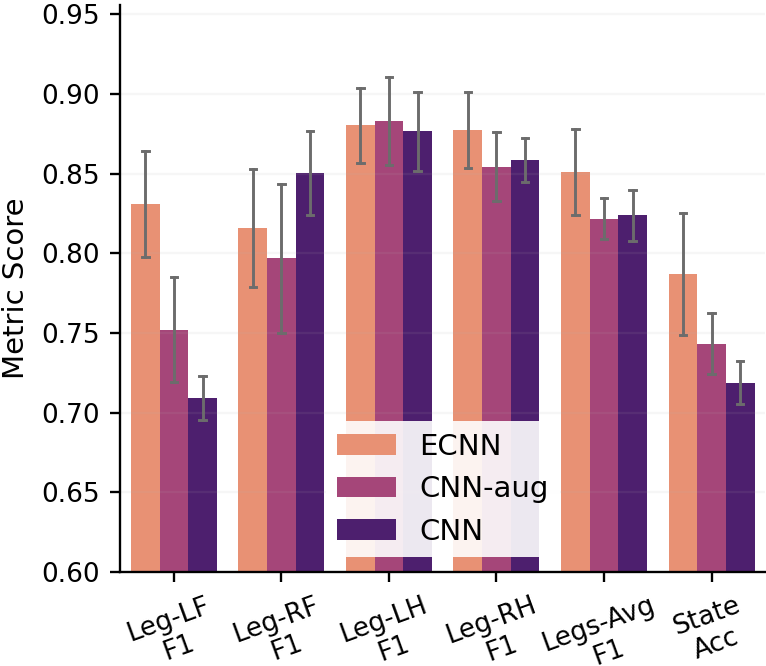}
            \end{subfigure}
            \caption{\textbf{Static-Friction-Regime contact detection results comparing CNN, CNN-aug, and ECNN.} \textbf{Left:} Sample efficiency in log-log scale. \textbf{Middle:} Average legs F1-score. \textbf{Right:} Classification metrics on test set performance of models trained with the entire training set. The selected metrics include contact-state ($\nnOut\in\R^{16}$) accuracy (Acc) and f1-score (F1) for each leg binary contact state. Due to the sagittal symmetry of the robot, the left front (LF) and right front (RF) legs are expected to be symmetric, as well as the left hind (LH) and right hind (RH) legs. F1-score is presented considering the dataset class imbalance (see \cref{sec:sup_contact_experiment_details}). The reported values represent the average and standard deviation across $8$ different seeds.
            \vspace{-.4cm}
            }
            \label{fig:contact_results}
        \end{figure*}
        In this experiment, we utilize the dataset introduced in \citet{lin2021deep_contact_estimation} for estimating static-friction-regime contacts in the foots of the Mini-Cheetah quadruped robot. The dataset consists of real-world proprioceptive data ($\qj$, $\dqj$, base linear acceleration, base angular velocity, and leg feet positions and velocities) captured over a history of $150$ time-frames. These measurements were obtained from inboard sensors during locomotion, encompassing various gaits and terrains. The dataset also includes $\nnOut\in\R^{16}$, representing the ground truth contact state of the robot, which was estimated offline using a non-causal algorithm. Our goal is to train a causal function approximator $\nn{\nnIn}$ to predict the contact state based on the input proprioceptive data.
        
        The Mini-Cheetah robot in the real-world exhibits an approximate reflection symmetry group, $\G\approx\Cyclic[2]$. As a result, both the proprioceptive data $\nnIn$ and the contact state $\nnOut$ share the symmetry group $\G$ (see \cref{sec:sup_contact_experiment_details}). In this experiment, we compare three variants of function approximators: the original Convolutional Neural Network architecture proposed by \citet{lin2021deep_contact_estimation} (CNN), a version of CNN trained with data augmentation (CNN-aug), and a version of CNN that incorporates hard-equivariance constraints (E-CNN).
        
        The sampling efficiency and average leg contact state classification results are depicted in \cref{fig:contact_results}-left-\&-middle. The equivariant model, E-CNN, demonstrates superior generalization performance and robustness to dataset biases compared to the unconstrained models (refer to \cref{sec:sup_contact_dataset}). Following E-CNN, CNN-aug exhibits better performance than the original CNN. In \cref{fig:contact_results}-right, we evaluate the classification metrics of the test set when using the entire training data. The E-CNN model outperforms both CNN-aug and CNN in contact state classification and average leg contact detection. Notably, exploiting symmetries helps mitigate suboptimal asymmetries in the models, preventing them from favoring the classification of one leg over others (observe legs LF and RF in \cref{fig:contact_results}-right). Further details can be found in \cref{sec:sup_contact_asymmetries}.
\section{Conclusions \& Discussion}
    \label{sec:conclusion}
    In this work, we present the definition of Discrete Morphological Symmetry (DMS). A capability of some dynamical systems to imitate the effect of rotations, translations, and infeasible reflections of space with a feasible discrete change in the system configuration. Using the language of group theory we study the set of DMSs of a dynamical system as a symmetry group $\G$ and conclude that: (1) A system with a symmetry group $\G$ exhibits $\G$-equivariant generalized mass matrix and dynamics. (2) That the symmetries of the dynamics extend to optimal control policies as well as to any proprioceptive and exteroceptive measurements, related to the evolution of the system's dynamics. 
    
    We establish the necessary theoretical abstractions to investigate and identify DMSs in any dynamical system, irrespective of the number of symmetries present. This new formalism allows us to identify the reflection/sagittal symmetry, prevalent in humans, animals, and most robots, as the simplest morphological symmetry group $\G=\Cyclic[2]$. Crucially, we use the same formalism to identify and exploit DMSs in real-world robotic systems with a greater number of symmetries. 

    In addition, we provide an open-access repository that facilitates the efficient prototyping of $\G$-equivariant neural networks for exploiting DMS in various applications involving rigid-body dynamics, such as robotics, computer graphics, and computational biology. This repository includes a growing collection of symmetric dynamical systems, with their corresponding symmetry groups already identified. Furthermore, we present compelling empirical and theoretical evidence supporting the utilization of DMSs in data-driven applications through data augmentation and the adoption of $\G$-equivariant neural networks. Both symmetry exploitation techniques result in improved sample efficiency and generalization.
    
    \subsubsection*{Limitations} Our work makes two assumptions: (1) that the system symmetry group $\G$ is finite, and (2) that the symmetries of the system are exact. For details  see \cref{sec:sup_limitations}.

    \subsubsection*{Further work} For data-driven applications the benefits of DMSs suggest the computational design of symmetrical dynamical systems. While for control applications, the  $\G$-equivariance nature of the generalized mass matrix suggests research on the numerical implications of this previously unexploited constraint in optimal control.
    
    
\section*{Acknowledgments}
This work’s experiments were run at the Barcelona Supercomputing Center in collaboration with the HPAI group. This work is supported by the Spanish government with the project MoHuCo PID2020-120049RB-I00 and the ERA-Net Chistera project IPALM PCI2019-103386.
\bibliographystyle{plainnat}
\bibliography{references}

\begin{thebibliography}{29}
\providecommand{\natexlab}[1]{#1}
\providecommand{\url}[1]{\texttt{#1}}
\expandafter\ifx\csname urlstyle\endcsname\relax
  \providecommand{\doi}[1]{doi: #1}\else
  \providecommand{\doi}{doi: \begingroup \urlstyle{rm}\Url}\fi

\bibitem[Abdolhosseini et~al.(2019)Abdolhosseini, Ling, Xie, Peng, and Van~de
  Panne]{abdolhosseini2019symmetric_locomotion}
Farzad Abdolhosseini, Hung~Yu Ling, Zhaoming Xie, Xue~Bin Peng, and Michiel
  Van~de Panne.
\newblock On learning symmetric locomotion.
\newblock In \emph{Motion, Interaction and Games}, pages 1--10. 2019.

\bibitem[Bronstein et~al.(2021)Bronstein, Bruna, Cohen, and
  Veli{\v{c}}kovi{\'c}]{bronstein2021geometric}
Michael~M Bronstein, Joan Bruna, Taco Cohen, and Petar Veli{\v{c}}kovi{\'c}.
\newblock Geometric deep learning: Grids, groups, graphs, geodesics, and
  gauges.
\newblock \emph{arXiv preprint arXiv:2104.13478}, 2021.

\bibitem[Carpentier et~al.(2019)Carpentier, Saurel, Buondonno, Mirabel,
  Lamiraux, Stasse, and Mansard]{carpentier2019pinocchio}
Justin Carpentier, Guilhem Saurel, Gabriele Buondonno, Joseph Mirabel, Florent
  Lamiraux, Olivier Stasse, and Nicolas Mansard.
\newblock The pinocchio c++ library: A fast and flexible implementation of
  rigid body dynamics algorithms and their analytical derivatives.
\newblock In \emph{2019 IEEE/SICE International Symposium on System Integration
  (SII)}, pages 614--619. IEEE, 2019.

\bibitem[Carter(2021)]{carter2021visual}
Nathan Carter.
\newblock \emph{Visual group theory}, volume~32.
\newblock American Mathematical Soc., 2021.

\bibitem[Finzi et~al.(2021{\natexlab{a}})Finzi, Benton, and
  Wilson]{finzi2021residualPP}
Marc Finzi, Gregory Benton, and Andrew~G Wilson.
\newblock Residual pathway priors for soft equivariance constraints.
\newblock \emph{Advances in Neural Information Processing Systems},
  34:\penalty0 30037--30049, 2021{\natexlab{a}}.

\bibitem[Finzi et~al.(2021{\natexlab{b}})Finzi, Welling, and
  Wilson]{finzi2021practical}
Marc Finzi, Max Welling, and Andrew~Gordon Wilson.
\newblock A practical method for constructing equivariant multilayer
  perceptrons for arbitrary matrix groups.
\newblock In \emph{International Conference on Machine Learning}, pages
  3318--3328. PMLR, 2021{\natexlab{b}}.

\bibitem[Funk et~al.(2021)Funk, Schaff, Madan, Yoneda, De~Jesus, Watson,
  Gordon, Widmaier, Bauer, Srinivasa, et~al.]{funk2021trifinger}
Niklas Funk, Charles Schaff, Rishabh Madan, Takuma Yoneda, Julen~Urain
  De~Jesus, Joe Watson, Ethan~K Gordon, Felix Widmaier, Stefan Bauer,
  Siddhartha~S Srinivasa, et~al.
\newblock Benchmarking structured policies and policy optimization for
  real-world dexterous object manipulation.
\newblock \emph{IEEE Robotics and Automation Letters}, 7\penalty0 (1):\penalty0
  478--485, 2021.

\bibitem[Glorot and Bengio(2010)]{glorot2010understanding}
Xavier Glorot and Yoshua Bengio.
\newblock Understanding the difficulty of training deep feedforward neural
  networks.
\newblock In \emph{Proceedings of the thirteenth international conference on
  artificial intelligence and statistics}, pages 249--256. JMLR Workshop and
  Conference Proceedings, 2010.

\bibitem[Hamed and Grizzle(2013)]{hamed2013event_lef_right}
Kaveh~Akbari Hamed and Jessy~W Grizzle.
\newblock Event-based stabilization of periodic orbits for underactuated 3-d
  bipedal robots with left-right symmetry.
\newblock \emph{IEEE Transactions on Robotics}, 30\penalty0 (2):\penalty0
  365--381, 2013.

\bibitem[He et~al.(2015)He, Zhang, Ren, and Sun]{he2015keiming_delving}
Kaiming He, Xiangyu Zhang, Shaoqing Ren, and Jian Sun.
\newblock Delving deep into rectifiers: Surpassing human-level performance on
  imagenet classification.
\newblock In \emph{Proceedings of the IEEE international conference on computer
  vision}, pages 1026--1034, 2015.

\bibitem[Holl{\'o}(2017)]{hollo2017demystification_symmetries_in_nature}
G{\'a}bor Holl{\'o}.
\newblock Demystification of animal symmetry: Symmetry is a response to
  mechanical forces.
\newblock \emph{Biology Direct}, 12\penalty0 (1):\penalty0 1--18, 2017.

\bibitem[Klambauer et~al.(2017)Klambauer, Unterthiner, Mayr, and
  Hochreiter]{klambauer2017self_normalizing_nn}
G{\"u}nter Klambauer, Thomas Unterthiner, Andreas Mayr, and Sepp Hochreiter.
\newblock Self-normalizing neural networks.
\newblock \emph{Advances in neural information processing systems}, 30, 2017.

\bibitem[Lanczos(2020)]{lanczos2020variational_principles_mechanics}
Cornelius Lanczos.
\newblock \emph{The variational principles of mechanics}.
\newblock University of Toronto press, 2020.

\bibitem[Lin et~al.(2021)Lin, Zhang, Yu, and
  Ghaffari]{lin2021deep_contact_estimation}
Tzu-Yuan Lin, Ray Zhang, Justin Yu, and Maani Ghaffari.
\newblock Legged robot state estimation using invariant kalman filtering and
  learned contact events.
\newblock In \emph{5th Annual Conference on Robot Learning}, 2021.

\bibitem[Noether(1918)]{noether1918invariante}
Emmy Noether.
\newblock Invariante variationsprobleme, math-phys.
\newblock \emph{Klasse, pp235-257}, 1918.

\bibitem[Ordonez-Apraez et~al.(2022)Ordonez-Apraez, Agudo, Moreno-Noguer, and
  Martin]{ordonez2022adaptable}
Daniel Ordonez-Apraez, Antonio Agudo, Francesc Moreno-Noguer, and Mario Martin.
\newblock An adaptable approach to learn realistic legged locomotion without
  examples.
\newblock In \emph{2022 International Conference on Robotics and Automation
  (ICRA)}, pages 4671--4678. IEEE, 2022.

\bibitem[Orin et~al.(2013)Orin, Goswami, and
  Lee]{orin2013centroidal_momentum_matrix_cmm}
David~E Orin, Ambarish Goswami, and Sung-Hee Lee.
\newblock Centroidal dynamics of a humanoid robot.
\newblock \emph{Autonomous robots}, 35\penalty0 (2):\penalty0 161--176, 2013.

\bibitem[Ostrowski and Burdick(1996)]{ostrowski1996geometric_perspectives}
Jim Ostrowski and Joel Burdick.
\newblock Geometric perspectives on the mechanics and control of robotic
  locomotion.
\newblock In \emph{Robotics Research}, pages 536--547. Springer, 1996.

\bibitem[Quigley(1973)]{quigley1973pseudovectors}
Robert~J Quigley.
\newblock Pseudovectors and reflections.
\newblock \emph{American Journal of Physics}, 41\penalty0 (3):\penalty0
  428--430, 1973.

\bibitem[Ravanbakhsh et~al.(2017)Ravanbakhsh, Schneider, and
  Poczos]{ravanbakhsh2017equivariance_parameter_sharing}
Siamak Ravanbakhsh, Jeff Schneider, and Barnabas Poczos.
\newblock Equivariance through parameter-sharing.
\newblock In \emph{International conference on machine learning}, pages
  2892--2901. PMLR, 2017.

\bibitem[Selig(2005)]{selig2005geometric_fundamentals_robotics}
Jon~M Selig.
\newblock \emph{Geometric fundamentals of robotics}, volume 128.
\newblock Springer, 2005.

\bibitem[Simpkins(2012)]{simpkins2012systemID}
Alex Simpkins.
\newblock System identification: Theory for the user, (ljung, l.; 1999)[on the
  shelf].
\newblock \emph{IEEE Robotics \& Automation Magazine}, 19\penalty0
  (2):\penalty0 95--96, 2012.

\bibitem[Van~der Pol et~al.(2020)Van~der Pol, Worrall, van Hoof, Oliehoek, and
  Welling]{van2020mdp}
Elise Van~der Pol, Daniel Worrall, Herke van Hoof, Frans Oliehoek, and Max
  Welling.
\newblock Mdp homomorphic networks: Group symmetries in reinforcement learning.
\newblock \emph{Advances in Neural Information Processing Systems},
  33:\penalty0 4199--4210, 2020.

\bibitem[Wang et~al.(2022)Wang, Walters, and Yu]{wang2022approximately}
Rui Wang, Robin Walters, and Rose Yu.
\newblock Approximately equivariant networks for imperfectly symmetric
  dynamics.
\newblock \emph{arXiv preprint arXiv:2201.11969}, 2022.

\bibitem[Wheeler(2014)]{wheeler2014covariance_eom}
James~T. Wheeler.
\newblock General coordinate covariance of the euler lagrange equations.
\newblock Classical Mechanics class notes, 2014.
\newblock URL
  \url{http://www.physics.usu.edu/Wheeler/ClassicalMechanics/CMCoordinateinvarianceofEulerLagrange.pdf}.

\bibitem[Wieber(2006)]{wieber2006holonomy}
P-B Wieber.
\newblock Holonomy and nonholonomy in the dynamics of articulated motion.
\newblock In \emph{Fast motions in biomechanics and robotics}, pages 411--425.
  Springer, 2006.

\bibitem[Yeh et~al.(2019)Yeh, Hu, and Schwing]{yeh2019chirality}
Raymond Yeh, Yuan-Ting Hu, and Alexander Schwing.
\newblock Chirality nets for human pose regression.
\newblock \emph{Advances in Neural Information Processing Systems}, 32, 2019.

\bibitem[Yu et~al.(2018)Yu, Turk, and Liu]{yu2018learning}
Wenhao Yu, Greg Turk, and C~Karen Liu.
\newblock Learning symmetric and low-energy locomotion.
\newblock \emph{ACM Transactions on Graphics (TOG)}, 37\penalty0 (4):\penalty0
  1--12, 2018.

\bibitem[Zinkevich and Balch(2001)]{zinkevich2001symmetry_mdp_implications}
Martin Zinkevich and Tucker Balch.
\newblock Symmetry in markov decision processes and its implications for single
  agent and multi agent learning.
\newblock In \emph{In Proceedings of the 18th International Conference on
  Machine Learning}. Citeseer, 2001.

\end{thebibliography}

\newpage
{
    
    \label{sec:supplementary}

    \appendices 
    \columnbreak
    \section{Properties of Robotic Systems with DMSs}
    \label{sec:sup_properties_morphological_symmetries}
            \begin{figure*}[!t]
                \centering
                \input{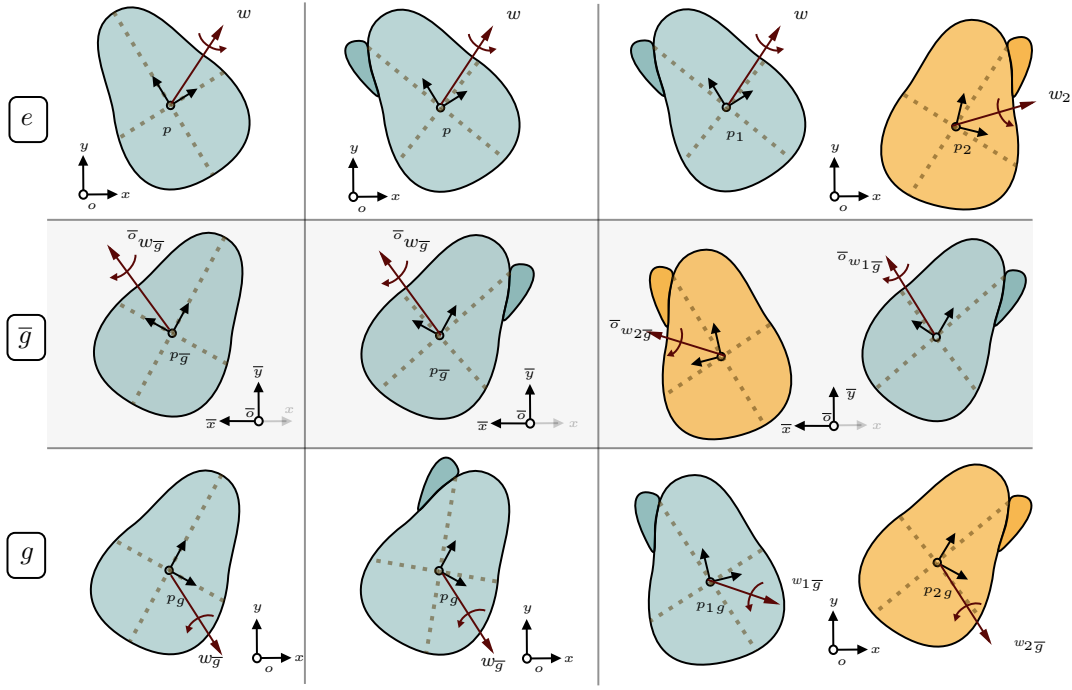}
                \caption{Properties of bodies capable of imitating a true reflection $\impg$ of space (w.r.t the $yz$-plane in this case), with a proper transformation $\g$ involving only a rotation and translation. The first row shows the original bodies with their respective angular velocities $\angvel$, subjected to trivial symmetry transformation $e$ (dashed lines represent the principle axes of inertia of the bodies), and the second and third rows display the effect of $\impg$ and $\g$ on the bodies and angular velocities, respectively. The first column displays a rigid body with symmetric mass distribution, for which $\g$ exists, as the reflected and rotated bodies share an equivalent angular kinetic energy. The second column shows a rigid body with asymmetrical mass distribution, for which the rotation $\g$, that produces a kinematic symmetry, results in the reflected and rotated bodies having different angular kinetic energies (\cref{eq:lagrangian-g_invariance}). The third column shows two bodies with asymmetrical mass distributions, each a reflected version of the other, in this case, the action $\g$ swaps bodies configurations to imitate the configuration and energy state of the reflected bodies transformed with $\impg$. Angular velocity is a pseudo-vector (or axial-vector), for which a reflection transformation is computed as $\angvel[\impg] = |\Rotation[\impg]|\Rotation[\impg]\angvel$ (see \citet{quigley1973pseudovectors}).
                \vspace{-0.4cm}}
                \label{fig:sup_rigid_body_symmetry_plane}
            \end{figure*}
        Here, we present a geometric (instead of albegraic) development analog to \cref{sec:dms_rigid_body_dynamics}. For clarity of the explanation, let us imagine two different Euclidean spaces and two versions of the robot: the original space (with reference frame $\world$) and robot with coordinates $\q$ and $\dq$, and the virtual rotated/reflected space (with a reference frame 
        $\refworld$, with configuration 
            $
                \refFrom{\world}{\memSE[\refworld]} = 
                \begin{bsmallmatrix} 
                    \Rotation[\impg] & \pos[\refworld] \\
                    \boldsymbol{0} & 1
                \end{bsmallmatrix}
            $
        ) and virtual robot with coordinates $\gq[\impg]$ and $\gdq[\impg]$ referenced to $\refworld$. Noting that in the case of a reflection, the virtual robot has reflected versions of each rigid body. 
        
        For \cref{eq:discrete_morphological_symmetries,eq:eom_g-equivariance} to hold, there must exist an action $\g \in \G$ transforming the real robot configuration $\gq, \gdq$ resulting in the same kinetic energy as the virtual robot's kinetic energy:%
        \begin{equation}
            \begin{split}
                {\KinE[\gq, \gdq]} 
                &= 
                \tfrac{1}{2} 
                \sum_{\bodyIdx=1}^{\nb}
                    \mass[\bodyIdx] {\gvel[\bodyIdx]}^{\transpose} {\gvel[\bodyIdx]} 
                    +
                    {\gangvel[\bodyIdx]^{\transpose}} {{\Inertia[\bodyIdx]}} {\gangvel[\bodyIdx]} 
                \\
                &\doteq
                \tfrac{1}{2} 
                \sum_{\impgBodyIdx=1}^{\nb}
                    \refmass[\impgBodyIdx] {\refvel[\impgBodyIdx]}^{\transpose} {\refvel[\impgBodyIdx]} 
                    +
                    {\refangvel[\impgBodyIdx]^{\transpose}} {{\RefInertia[\impgBodyIdx]}} {\refangvel[\impgBodyIdx]} 
                =
                {\KinE[{\gq[\impg], \gdq[\impg]}]},
            \end{split}
            \label{eq:kin_energy}
        \end{equation}
        where $\gvel[i]$, $\gangvel[i]$, $\mass[i]$ and $\Inertia[i]$ are the linear and angular velocity, mass, and inertia matrix of the transformed body $i$ (referenced to $\world$). Likewise, $\refvel[i]$, $\refangvel[i]$, $\refmass[i]$ and $\RefInertia[i]$ are the equivalent quantities for the virtual robot and body $i$ (referenced to $\refworld$).
        
        \subsection{Symmetries of kinematic parameters:} 
        \label{sec:sup_kinematic_symmetries}
            Ignore momentarily the influence of the mass and inertia in terms of the real and virtual bodies. We can assert that for \cref{eq:kin_energy} to hold, the transformed configuration should result in a kinematic tree indistinguishable from the virtual robot's. Thus, for everybody $\bodyIdx$ in the real robot kinematic tree, there should exist an equivalent virtual body $\impgBodyIdx$ (as seen in \cref{fig:examples_morphological_symmetries}, not always $\impgBodyIdx=\bodyIdx$). By equating the linear and angular velocities of the real and virtual bodies, referenced to $\world$,
            and expressing the velocities as functions of the generalized coordinates we obtain:
            \begin{align}
                    \gvel[\bodyIdx] &= 
                        \refvel[\impgBodyIdx] \justification{\doteq
                            \Rotation[\impg] \cdot \vel[\impgBodyIdx]
                        }
                    \nonumber
                    \\
                    \PosJacob[\bodyIdx](\gq) \gdq &= 
                        \Rotation[\impg] \cdot \PosJacob[\impgBodyIdx](\q) \dq 
                    \nonumber
                    \\
                    \PosJacob[\bodyIdx](\gq) \g &= 
                        \Rotation[\impg] \cdot \PosJacob[\impgBodyIdx](\q) 
                    \\
                    \gangvel[\bodyIdx] &=
                        \refangvel[\impgBodyIdx] \justification{\doteq
                            |\Rotation[\impg]|\Rotation[\impg] \cdot \angvel[\impgBodyIdx]
                        }
                    \nonumber
                    \\
                    \OriJacob[\bodyIdx](\gq) \gdq &=
                        |\Rotation[\impg]|\Rotation[\impg] \cdot \OriJacob[\impgBodyIdx](\q) \dq,
                    \nonumber
                    \\          
                    \OriJacob[\bodyIdx](\gq) \g &=
                        |\Rotation[\impg]|\Rotation[\impg] \cdot \OriJacob[\impgBodyIdx](\q),
                \label{eq:sup_kinematic_symemtries}
            \end{align}
            where $\PosJacob[i](\q), \; \OriJacob[i](\q) \in \R^{3 \times \nq}$ are the position and orientation analytical Jacobians (describing the instantaneous velocity vectors contributed by each DoF to body $i$) of the real robot at a configuration $\q$ \citep{wieber2006holonomy}. Formulating \cref{eq:sup_kinematic_symemtries} for each of the $\nb$ bodies of the robot we obtain at best $\nb \times 3 \times \nq$ non-linear equations that can be used to assert if $\g$ exists. In practice, the action representation $\rep[\ConfSpace]{\g}$ and especially its component acting on the joint space $\rep[\ConfSpaceJS]{\g}$ can be trivially determined by solving \cref{eq:sup_kinematic_symemtries} (or equivalently \cref{eq:kinematic_symmetries_rigid_bodies}) for each body from top to bottom of the kinematic tree (i.e., base first, end-effectors last), if $\g$ exists.
        \subsection{Symmetries of dynamic prameters}
        \label{sec:sup_reflection_symmetry}
            Let us assume kinematic symmetry and direct our attention now to the influence of the mass and inertia terms on the kinetic energy of a single rigid body when it is transformed with the action $g$, which imitates a true reflection of space $\impg$. Focus on the first two columns of \cref{fig:sup_rigid_body_symmetry_plane}. Because of the kinematic symmetry the CoM of the reflected and transformed bodies coincide, both bodies have equivalent linear components of kinetic energy. However, for arbitrary rigid bodies, the reflected and transform bodies will have different angular components of kinetic energy. Note that in the general case, the transformed and reflected bodies' inertia will differ, thus even if both bodies have the same angular velocities, their kinetic energy will differ. 
    
            Let $\PrincipalFrame$, $\PrincipalFrame[\impg]$ and $\PrincipalFrame[\g]$ be frames located at the CoM of the original, reflected and transformed bodies, aligned with the principal axes of inertia of each of the bodies. Similarly, denote $\refFrom{\world}{\Inertia}$ and $\refFrom{\world}{\Inertia[\g]}$ 
            as the original and transformed bodies inertias referenced to $\world$, and $\refFrom{\refworld}{\RefInertia[\impg]}$ as the reflected body inertia referenced to the reflected Euclidean space $\refworld$. In order to comply with \cref{eq:sup_kinematic_symemtries}, we must ensure that:
            \begin{subequations}
                \begin{alignat}{2}
                    \refToFrom{^\world}{\angvel}{^\transpose_{\impg}} \refFrom{\world}{\Inertia[\g]} \refToFrom{^\world}{\angvel}{_{\impg}}
                    &=
                    \refToFrom{^\refworld}{\angvel}{^\transpose_{\impg}} \refFrom{\refworld}{\RefInertia[\impg]} \refToFrom{^\refworld}{\angvel}{_{\impg}},
                    ,
                    \nonumber
                    \\
                    (\Rotation[\impg]
                    \refToFrom{^\world}{\angvel}{})^\transpose \refFrom{\world}{\Inertia[\g]} (\Rotation[\impg]
                    \refToFrom{^\world}{\angvel}{})
                    &=
                    \refToFrom{^\refworld}{\angvel}{^\transpose_{\impg}} \refFrom{\refworld}{\RefInertia[\impg]} \refToFrom{^\refworld}{\angvel}{_{\impg}}
                    \quad 
                    \nonumber
                    \\
                    &\qquad |\;
                    \justification{
                        \refFrom{\world}{\angvel[\impg]} = |\Rotation[\impg]|\Rotation[\impg]\refFrom{\world}{\angvel}
                    }
                    ,
                    \nonumber
                    \\
                    \refFrom{\world}{\Inertia[\g]} 
                    &= 
                    \Rotation[\impg] 
                        \refFrom{\refworld}{\RefInertia[\impg]}
                    \Rotation[\impg]
                    \nonumber
                    \\
                    &\qquad |\;
                    \justification{
                       \refToFrom{^\world}{\angvel}{} \equiv \refToFrom{^\refworld}{\angvel}{_{\impg}}
                       , \quad
                       \Rotation[\impg] \Rotation[\impg] = \Identity
                    }
                    ,
                    \nonumber
                    \\
                    \refToFrom{^\world}{\Rotation}{^{\PrincipalFrame[\g]}}
                        \refFrom{\PrincipalFrame[\g]}{\Inertia[\g]} 
                    \refToFrom{^\world}{\Rotation}{^{\PrincipalFrame[\g]}}^\transpose
                    &= 
                    \Rotation[\impg] 
                    \refToFrom{^\refworld}{\Rotation}{^{\PrincipalFrame[\impg]}}
                        \refFrom{\PrincipalFrame[\impg]}{\RefInertia[\impg]}
                    \refToFrom{^\refworld}{\Rotation}{^{\PrincipalFrame[\impg]}}^\transpose
                    \Rotation[\impg]
                    \nonumber
                    \\
                    &\qquad |\;
                    \justification{
                        \refFrom{a}{\Inertia} = \refToFrom{^b}{\Rotation}{^a}\refFrom{b}{\Inertia}\refToFrom{^b}{\Rotation}{^a}^\transpose
                        }
                    ,
                    \nonumber
                    \\
                    \refToFrom{^\world}{\Rotation}{^{\PrincipalFrame[\g]}}
                        \refFrom{\PrincipalFrame[\g]}{\Inertia[\g]} 
                    \refToFrom{^\world}{\Rotation}{^{\PrincipalFrame[\g]}}^\transpose
                    &= 
                    \Rotation[\impg] 
                    \refToFrom{^\world}{\Rotation}{^\PrincipalFrame}
                    \refToFrom{^\PrincipalFrame}{\Rotation}{^{\PrincipalFrame[\impg]}}
                        \refFrom{\PrincipalFrame[\impg]}{\RefInertia[\impg]}
                    \refToFrom{^\PrincipalFrame}{\Rotation}{^{\PrincipalFrame[\impg]}}^\transpose
                    \refToFrom{^\refworld}{\Rotation}{^{\PrincipalFrame[\impg]}}^\transpose
                    \Rotation[\impg]
                    ,
                    \nonumber
                    \\
                    \refToFrom{^\world}{\Rotation}{^{\PrincipalFrame[\g]}} 
                    &\doteq \Rotation[\impg] \refToFrom{^\world}{\Rotation}{^\PrincipalFrame}
                    \refToFrom{^\PrincipalFrame}{\Rotation}{^{\PrincipalFrame[\impg]}}
                    \label{eq:symmetric_rigid_bodies-b}
                    \\
                    &\qquad |\;
                    \justification{
                        \refFrom{\PrincipalFrame}{\Inertia} \equiv 
                        \refFrom{\PrincipalFrame[\g]}{\Inertia[\g]} \equiv
                        \refFrom{\PrincipalFrame[\impg]}{\RefInertia[\impg]}
                    }.
                    \nonumber
                \end{alignat}
                \label{eq:symmetric_rigid_bodies}
            \end{subequations}
            What \cref{eq:symmetric_rigid_bodies-b} states is that in order for the reflected and transformed bodies to have equivalent angular kinetic energy, both bodies should have co-linear (or aligned) principal axes of inertia. This allows us to describe $\refToFrom{^\world}{\Rotation}{^{\PrincipalFrame[\g]}}$ as a function of the original body configuration $\refToFrom{^\world}{\Rotation}{^{\PrincipalFrame}}$ and two reflection matrices: the true reflection of space $\Rotation[\impg]$ and a body specific diagonal reflection matrix $\refToFrom{^\PrincipalFrame}{\Rotation}{^{\PrincipalFrame[\impg]}}$, which exists only if the rigid body has a symmetric mass distribution. A visual example for symmetric and asymmetrical rigid bodies is presented in \cref{fig:sup_rigid_body_symmetry_plane} left and middle columns. A similar analysis follows for DMSs imitating rotations
        \subsection{Symmetric position and velocity constraint configuration spaces}
            Although it is implicitly implied on \cref{eq:lagrangian-g_invariance} that the constrained position $\ConfSpace$ and velocity $\TangConfSpace$ configuration vector spaces should also be symmetric or equivariant, this property might be easily overlooked. As mentioned in \cref{sec:discrete_morphological_symmetries} the relevance of morphological symmetries relies on the equivariant nature of the system dynamics (\cref{eq:eom_g-equivariance,eq:equivariance-invariance-constraints}), which imprints symmetry constraints on optimal control policies and proprioceptive and exteroceptive measurements. However, with non-symmetric constrained configuration spaces, \cref{eq:lagrangian-g_invariance} will not hold for every system state $\q \in \ConfSpace$, $\dq \in \TangConfSpace$, and any uncontrolled or controlled trajectory of the system dynamics shall not have a symmetric equivalent trajectory, as this has the potential to violate the constraints of the configuration space. 
    
    
    \section{Efficient Construction of $\G$-equivariant NNs for DMS Groups $\G$}
    \label{sec:sup_ml_contributions}
    
        As mentioned in \cref{sec:machine_learning_symmetries} our work builds upon the framework for the construction of $\G$-equivariant NN of \citet{finzi2021practical}. The core limitation of this framework is the inability to handle large dimensional spaces, due to the computational and memory complexities. For instance, for an equivariant layer with input dimension $n$ and output dimension $m$,  the computational complexity of finding the equivariant linear map basis $\basisEquivT$ (which is quadratic $\mathcal{O}((mn)^2r^2)$ through the Krylov subspace method) and the memory complexity of $\basisEquivT \in \R^{mn \times r} \;|\; r \leq mn$, become easily intractable for moderate $n$ and $m$ dimensions. This limitation is openly discussed in the EMLP repository \textit{README.md}, but regretfully not in the original paper.
        
        In practice, we found these limitations when trying to construct the equivariant version of the Contact CNN \citep{lin2021deep_contact_estimation} in our second experiment. This architecture in its internal layers has $n,m > 2000$, for which: (i) the Krylov subspace method complexity renders the operation intractable with standard hardware and (ii) the matrices $\basisEquivT$ of internal layers required storage of $1[Gb]>$ for moderate input output dimensions ($m,n \approx 250$) and $1[Pb]>$ for $m,n > 2000$). See \cref{table:cnn-ecnn_comparison} for a comparison between dense and sparse matrix representations. 
    \subsection{Trainable parameter reduction of $\G$-equivariant layers (for $\G$ a DMS group)}
    \label{sec:sup_parameter_reduction}
         Determining analytically the number of trainable parameters (i.e. the rank $\basisRank$) of an $\G$-equivariant layer is, in general, an unresolved problem. However, for DMS groups, $\basisRank$ can be computed once the input-output action representations are known. The requirement to compute $\basisRank$ is that actions affecting the linear maps are a semi-direct product\footref{foot:semi-direct-product} of the input-output groups, and the input-output representations are generalized permutation matrices. These conditions are met for most DMS groups (see \cref{sec:sup_limitations}).
    
        The equivariance constraints of \cref{eq:equivariant_linear_maps_basis} on linear maps of perceptron (or convolutional) layers imply a reduction of trainable parameters from $|\nnWflat|=mn$ to $|\basisCoefV|=\basisRank \leq mn$. For DMS groups, $\basisRank$ is associated with the number of unique orbits of the elements of $\nnWflat$. Thus we can compute this value using the orbit-counting theorem (also known as Burnside's Lemma), which states that the number of orbits is the average number of fix-points of $\G$, that is $\basisRank = \tfrac{1}{|\G|} \sum_{g \in \G} |\nnWflat^{\g}|$, 
            where 
            $\nnWflat^{\g} \doteq \{\nnWflatEV \in \nnWflat: \g \cdot \nnWflatEV = \nnWflatEV \}$ 
            represents the set of elements of $\nnWflat$ that are invariant to $\g$ (i.e. fix-points). Those fix-points can be identified by the elements on the diagonal of $\rep[\nnW]{\g}$ that are equal to one. Therefore, for a $\G$-equivariant layer, the number of trainable parameters is determined by:
            \begin{equation}
                \basisRank = \tfrac{1}{|\G|} {\textstyle \sum}_{g \in \G} \invtrace[\rho_{\nnW}]{\g} = \tfrac{1}{|\G|} {\textstyle \sum}_{g \in \G} \; \invtrace[\rho_{in}]{\g^{\text{-}1}} \cdot \invtrace[\rho_{out}]{\g}, 
                \label{eq:number_of_trainable_params}
            \end{equation}
            denoting $\invtrace[\rho]{\g} : \G \rightarrow \N$ as the number of fix-points of the action representation $\rep{g}$. Therefore, the number of trainable parameters can range from $\sfrac{|\nnWflat|}{|\G|} \leq r \leq |\nnWflat|$, depending on the fix-points of the layers' input and output spaces.
        
    \subsection{Parameter initialization of equivariant layers for DMS}
    \label{sec:sup_initialization_equivariant_layers}
        
        Consider a Equivariant Neural Network architecture composed of multiple layers of equivariant linear (or convolutional) layers of the form $^{l}\nnOut : = \nnAct(\,^{l}\nnW  \,^{l}\nnIn + \,^{l}\nnBias)$, being $l$ the layer index, $^{l}\nnIn\in \R^{n}$ and $^{l}\nnOut\in \R^{m}$ the layer's input and output vector spaces, $^{l}\nnW \doteq {\textstyle \sum_{k}^{r}} \,^{l}\basisCoefEV[k]\,^{l}\basisEquivT[:,:,k] \in \R^{m \times n}$ the layer's linear map, $\,^{l}\basisEquivT \in \R^{m \times n \times \basisRank}$ the layer's $\basisRank$ basis vectors spawning the space of equivariant linear maps, $\,^{l}\basisCoefV \in \R^{\basisRank}$ the layer's trainable parameters, and $^{l}\nnBias \in \R^{m}$ the layer's bias vector.
        
        For the optimal flow of information throughout the network, it's relevant to initialize the trainable parameters such that the variance of activations (during inference/forward-propagation) and gradients (during back-propagation) is kept constant, avoiding activations/gradients from vanishing or exploiting \citep{glorot2010understanding}\footnote{See Pierre Ouannes blog: \href{https:\/\/pouannes.github.io\/blog\/initialization\/\#xavier-and-kaiming-initialization}{pouannes.github.io/blog/initialization}}. 
        
        The derivation is based on the equivalent process for unconstrained layers presented in \citet{he2015keiming_delving}. Let the layer's activations before the non-linearity be denoted by $^{l}\vz = \,^{l}\nnW  \,^{l}\nnIn + \,^{l}\nnBias$, such that $^{l}\nnOut = \nnAct(^{l}\vz)$, and note that $\,^{l}\nnIn = \,^{l-1}\nnOut$. Furthermore, we will assume the elements of $^{l}\basisCoefV$ and $^{l}\nnIn$ are mutually independent and sampled from two independent distributions, denoting the random variables of the two distributions as $^{l}\basisCoefEV$ and $^{l}\nnInRand$.
        
        The core difference in the initialization of unconstrained and equivariant layers lies in the way the linear map is parameterized. For equivariant layers we have: 
        \begin{align}    
            \Var(\,^{l}\nnW  \,^{l}\nnIn + \,^{l}\nnBias) 
            &= 
           \sum_{i}^{m}\sum_{j}^{n}\Var\left(\,^{l}\nnWEV[i,j]  \,^{l}\nnInRand[j]\right)
               |\; {\justification{\Var(\,^{l}\nnBias) = 0}}
            \nonumber
            \\
            &= 
            \sum_{i}^{m}\sum_{j}^{n}\Var\left(\left(\sum_{k}^{\basisRank} \,^{l}\basisCoefEV[k]\,^{l}\basisEquivT[m,n,k] \right)  \,^{l}\nnInRand[j]\right)
            \nonumber
            \\,
            &= \Var\left(\,^{l}\basisCoefEV\,^{l}\nnInRand\right) 
                \ubcolor{awesomeblue}{
                    \sum_{i}^{m}\sum_{j}^{n}\sum_{k}^{\basisRank}\basisEquivT[m,n,k]^2
                }{\lambda_{^{l}\basisEquivT}}
                \label{eq:variance_equiv_basis}
                \\ \nonumber
                &\qquad |\;
                \justification{\small
                    \Var\left( \sum_{a} \underbracket{s_{a}}_{const}\rvp \right) = \sum_{a}s_{a}^2\Var(\rvp)
                }
        \end{align}  
        In the forward-propagation scenario, we are interested in conserving the variance of the activations throughout layers, that is we must ensure $\Var(\,^{l}\rvz) = \Var(\,^{l-1}\rvz)$. Using \cref{eq:variance_equiv_basis} we obtain:
        \begin{align}    
            \Var(\,^{l}\vz) 
            &= 
            \Var(\,^{l}\nnW  \,^{l}\nnIn + \,^{l}\nnBias) 
            \nonumber
            \\
            m \; \Var(\,^{l}\rvz) 
            &= \lambda_{^{l}\basisEquivT} \Var\left(\,^{l}\basisCoefEV\,^{l}\nnInRand\right) 
            \nonumber
            \\
            \Var(\,^{l}\rvz) 
            &= \frac{\lambda_{^{l}\basisEquivT}}{m} 
                \left( 
                    \ubcolor{awesomeblue}{\E(^{l}\basisCoefEV^2)}{\Var(^{l}\basisCoefEV)} \E(\,^{l}\nnInRand^2) 
                        - 
                    \ubcolor{awesomeblue}{\E(^{l}\basisCoefEV)^2}{=0} \E(\,^{l}\nnInRand)^2 
                \right) 
            \nonumber
            \\
            \Var(\,^{l}\rvz) 
            &= \frac{\lambda_{^{l}\basisEquivT}}{m}  
                \Var(^{l}\basisCoefEV) \E\left(\,^{l-1}\nnOutRand^2\right)
                \nonumber\\
                &\qquad \;|\; \justification{^{l}\nnInRand =\,^{l-1}\nnOutRand = \nnAct(\,^{l-1}\rvz)}
            \nonumber
            \\
            \Var(\,^{l}\rvz) 
            &= \frac{\lambda_{^{l}\basisEquivT} \lambda_{\nnAct}}{m} 
                \Var(^{l}\basisCoefEV)\Var(\,^{l-1}\rvz) 
            \label{eq:sup_forward_prop_case_nonlinearity}
                \\ \nonumber
                &\qquad \;|\; \justification{\E\left(\,^{l-1}\nnOutRand^2\right) = \lambda_{\nnAct}\Var\left(\,^{l-1}\rvz\right)
                }
            \\
            \Var(\,^{l}\rvz) 
            &\equiv \Var(\,^{l-1}\rvz) 
            \label{eq:sup_forward_prop_case_variance}
                \\ \nonumber
                &\qquad \;|\; \justification{\Var\left(^{l}\basisCoefEV\right) \doteq \frac{m}{\lambda_{^{l}\basisEquivT} \lambda_{\nnAct}}} 
        \end{align}  
        where $\lambda_{\nnAct}$ in \cref{eq:sup_forward_prop_case_nonlinearity} is a non-linearity dependent scalar computed analytically or empirically (see \citet{he2015keiming_delving}). In \cref{eq:sup_forward_prop_case_variance} we conclude that if we sample the equivariant layer trainable parameters $^{l}\basisCoefV$ from a distribution ensuring $\Var(^{l}\basisCoefEV) \doteq \frac{m}{\lambda_{^{l}\basisEquivT} \lambda_{\nnAct}}$, the variance of the activations across equivariant layers remain constant. A similar procedure can be applied to the backward propagation case, concluding that in order to maintain a constant variance of the gradients across the network layers we should sample the trainable parameters ensuring $\Var(^{l}\basisCoefEV) \doteq \frac{n}{\lambda_{^{l}\basisEquivT} \lambda_{\nnAct}}$. As remarked in \citet{he2015keiming_delving} both variance values for the forward and backward propagation cases lead to the proper flow of information in the network. On \cref{fig:sup_initialization}, it can be appreciated that our method achieves equivalent results for equivariant architectures as \cite{he2015keiming_delving} does for standard linear and convolutional architectures.
        %
    
    \section{Implementation Details \& Code}
    \label{sec:sup_experiment_details}
        \begin{table*}[!t]
            \label{table:cnn-ecnn_comparison}
            \begin{center}
            \begin{tabular}{ c c c | r r | r r |}
                & & &  \multicolumn{2}{c}{Dense Memory [Bytes]} & \multicolumn{2}{c}{Sparse Memory [Bytes]} 
                \\
                \textbf{Layer Type} & $n$ & $m$ & $\rep[\nnWflat]{g}$ & $\basisEquivT$ & $\rep[\nnWflat]{g}$ & $\basisEquivT$
                \\
                \hline
                1D-Conv & $ 54   $ & $ 64   $  & $ 764.41 M $ & $ 191.10 M $ & $ 221.18 k $  & $ 110.59 k $  \\ 
                1D-Conv & $ 64   $ & $ 64   $  & $ 1.07   G $ & $ 268.43 M $ & $ 262.14 k $  & $ 131.07 k $  \\ 
                1D-Conv & $ 64   $ & $ 128  $  & $ 4.29   G $ & $ 1.07   G $ & $ 524.28 k $  & $ 262.14 k $  \\ 
                1D-Conv & $ 128  $ & $ 128  $  & $ 17.18  G $ & $ 4.29   G $ & $   1.04 M $  & $ 524.28 k $  \\ 
                Percept & $ 4736 $ & $ 2048 $  & $ 6.02   P $ & $ 1.50   P $ & $ 620.75 M $  & $ 310.37 M $  \\ 
                Percept & $ 2048 $ & $ 512  $  & $ 70.36  T $ & $ 17.59  T $ & $  67.10 M $  & $  33.55 M $  \\ 
                Percept & $ 512  $ & $ 16   $  & $ 4.29   G $ & $ 1.07   G $ & $ 524.28 k $  & $ 262.14 k $  \\ 
            \end{tabular}
            \end{center}
            \caption{
                \textbf{Comparison of memory complexity of individual layers of the equivariant version of Contact-CNN \cite{lin2021deep_contact_estimation} (ECNN)}. This example compares the sparse and dense representations of matrices $\basisEquivT \in \R^{mn \times r}$ and the $|\G|$ group action representations $\rep[\nnWflat]{g} \in \R^{mn \times mn}$, for the symmetry group $\G=\Cyclic[2]$ of the Mini-Cheetah robot, with $\basisRank = \sfrac{mn}{2}$ (see \cref{eq:number_of_trainable_params}). Here, $n,m$ represents the input and output dimensions of each layer.  The dense memory complexity of all action representations increases with the group order $|\G|$ while the memory complexity for $\basisEquivT$ decreases with larger group orders (since $\basisRank \leq mn$ becomes smaller). We assume floating point representations with $32$ bits.
            }
        \end{table*}
        Additional to this section, we provide open-access code with the scripts for reproducing the experiments of this work, the parameters of the models used for comparison, along with additional interactive examples visualizing morphological symmetries of both robotic systems and data.
        
        \subsection{Efficient data augmentation}
        
        Since any input $\nnIn$ and output $\nnOut$ spaces of equivariant architectures have matrix symmetry action representations, $\rep[\nnInSpace]{\g}$, $\rep[\nnOutSpace]{\g}$, it is possible to perform batched data augmentation, reducing the computational complexity of augmenting a batch of $N_b$ samples from $N_b$ matrix-vector multiplications to a single matrix-matrix multiplication, preferably performed after data is loaded to GPU for optimal performance. 
        
        \subsection{Hyperparameter tunning}
        
        The only hyper-parameter tunned for each model and model variant was the learning rate. For all model variants presented in this work (except the original Contact-CNN model from \citet{lin2021deep_contact_estimation}, which we retrained using the same hyperparameters reported by the authors) we ran a grid-search in log-scale among 20 different learning rates. In this scenario, we always used the entire training dataset and optimized w.r.t computed loss in the entire validation partition. The learning rate values used for each model are depicted in \cref{table:hps}.
        \subsection{Experiment: CoM Momentum Estimation}
        \label{sec:sup_com_experiment_details}
        The dataset for the CoM estimation experiment was generated using Pinocchio \citep{carpentier2019pinocchio}, which in turn uses the URDF models of the robots Solo and Atlas, to extract the kinematic and dynamic parameters required to compute the Centroidal Momentum Matrix $\CMM(\q)$ matrix \citep{orin2013centroidal_momentum_matrix_cmm}, with which computing the CoM momentum reduces to: 
        \begin{align}
            \g \cdot \momentum &= \CMM(\g \cdot\qj) \g \cdot\dqj      & \qquad&|\quad \forall \g \in \G. 
            \label{eq:com_momentum_pin}
            \\
            \g \cdot \momentum &\approx \nn{\g \cdot\qj, \g \cdot\dqj} & \qquad&|\quad \forall \g \in \G. 
            \label{eq:com_momentum_nn}
        \end{align}
        Where \cref{eq:com_momentum_pin} expresses the analytical $\G$-equivariant function to compute the CoM momentum. While \cref{eq:com_momentum_nn} is the approximation of this function by an $\G$-equivariant NN, with parameters $\nnParams$.
    
        \subsubsection{Determination of the input and output representations $\rep[\nnInSpace]{\g}, \rep[\nnOutSpace]{\g} \;|\; \g \in \G$}
        \label{sec:sup_id_reps_com}
            Both robots Solo and Atlas evolve in the Euclidean space of $3$-dimensions. Therefore their configuration space can be decoupled into $\ConfSpace \doteq \EG[3] \times \ConfSpaceJS$. After identifying their symmetry groups and their corresponding $\EG[3]$ and $\ConfSpaceJS$ representations ($\rep[{\EG[3]}]{\g}, \rep[{\ConfSpace}]{\g} \; |\;\forall\; \g \in \G$), identifying the representations of the input and output spaces of the NN function approximator (\cref{eq:com_momentum_nn}) becomes a trivial task considering that:
            \begin{multline}
                \ubcolor{awesomeblue}{
                    \begin{bmatrix}
                         \rep[\EG]{\impg} & \boldsymbol{0}\\
                         \boldsymbol{0}      & \rep[\EG]{\impg}
                     \end{bmatrix}
                 }{\rep[\nnOutSpace]{\g}}
                 \ubcolor{black}{
                    \begin{bmatrix}
                        \linMomentum \\ 
                        \angMomentum
                    \end{bmatrix}
                }{\nnOut}
                \approx  
                f\bigg(
                    \ubcolor{awesomeorange}{
                        \begin{bmatrix}
                             \rep[\ConfSpaceJS]{\g} & \boldsymbol{0}\\
                             \boldsymbol{0}      & \rep[\ConfSpaceJS]{\g}
                         \end{bmatrix} 
                    }{\rep[\nnInSpace]{\g}}
                    \ubcolor{black}{        
                        \begin{bmatrix}
                            \qj \\ 
                            \dqj
                        \end{bmatrix} 
                    }{\nnIn}
                \bigg) \\
                |\quad \forall \quad (\g,\,\impg) | \g \in \G, \impg \in \EG[3]
                \label{eq:com_momentum_pin_symm}
            \end{multline}
            Defining 
            $\nnIn=\begin{bsmallmatrix}
                \qj \\ \dqj
            \end{bsmallmatrix}\in \R^{2\nj}$ and $\nnOut=\momentum \in \R^{2\dimfiber} \equiv \R^{6}$. Note that by definition any improper transformation applied to a pseudo-vector (e.g. angular velocity/momentum, torques) is computed as $|\Rotation|\Rotation \cdot \angMomentum$. 
        
            \subsubsection{Practical details of the dataset generation}
            The URDF files of the robots Solo and Atlas are generated using XACRO scripts, which replicate the structure of limbs to their symmetric counterparts, making the dynamics of the robots in simulation exactly $\G$-equivariant. However, the algorithm for computing the CoM momentum from Pinocchio is numerically sensitive, resulting in the orbits of the momentum $\G \cdot \momentum$ deviating slightly from the theoretical orbits. Therefore to reduce numerical errors and ensure the theoretical equivariance of the data, we replace every target variable by the average of its orbit $\nnOut=\momentum \doteq \frac{1}{|G|} \sum \G \cdot \g^-1(\CMM\left(\rep[\ConfSpace]{\g} \qj \right) \rep[\ConfSpace]{\g} \dqj)
                    \quad|\quad \forall \quad \g \in \G$. %
        
        \subsection{Experiment: Static-Friction-Regime Contact Detection}
        \label{sec:sup_contact_experiment_details}
            \begin{table}[!t]
                \label{sample-table}
                \begin{center}
                \begin{tabular}{ c c c c | c || c | c c c c }
                    \textbf{RF} & \textbf{\color{awesomeblue}LF} & \textbf{RH} & \textbf{\color{awesomeblue}LH} & $\nnOut$ &  $\g \cdot \nnOut$ & \textbf{\color{awesomeblue}LF} & \textbf{RF} & \textbf{\color{awesomeblue}LH} & \textbf{RH} \\
                    0 & 0 & 0 & 0 &    0    &    0    & 0 & 0 & 0 & 0 \\
                    0 & 0 & 0 & 1 &    1    &    2    & 0 & 0 & 1 & 0 \\
                    0 & 0 & 1 & 0 &    2    &    1    & 0 & 0 & 0 & 1 \\
                    0 & 0 & 1 & 1 &    3    &    3    & 0 & 0 & 1 & 1 \\
                    0 & 1 & 0 & 0 &    4    &    8    & 1 & 0 & 0 & 0 \\
                    0 & 1 & 0 & 1 &    5    &    10   & 1 & 0 & 1 & 0 \\
                    0 & 1 & 1 & 0 &    6    &    9    & 1 & 0 & 0 & 1 \\
                    0 & 1 & 1 & 1 &    7    &    11   & 1 & 0 & 1 & 1 \\
                    1 & 0 & 0 & 0 &    8    &    4    & 0 & 1 & 0 & 0 \\
                    1 & 0 & 0 & 1 &    9    &    6    & 0 & 1 & 1 & 0 \\
                    1 & 0 & 1 & 0 &    10   &    5    & 0 & 1 & 0 & 1 \\
                    1 & 0 & 1 & 1 &    11   &    7    & 0 & 1 & 1 & 1 \\
                    1 & 1 & 0 & 0 &    12   &    12   & 1 & 1 & 0 & 0 \\
                    1 & 1 & 0 & 1 &    13   &    14   & 1 & 1 & 1 & 0 \\
                    1 & 1 & 1 & 0 &    14   &    13   & 1 & 1 & 0 & 1 \\
                    1 & 1 & 1 & 1 &    15   &    15   & 1 & 1 & 1 & 1 \\
                    \label{table:contact_state_symmetry}
                \end{tabular}
                \end{center}
                \caption{Symmetric contact state for Mini-Cheetah quadruped robot. Considering its  morphological symmetry group $\Cyclic[2]=\{e,\g\}$. Each leg binary contact state (LF: Left Front, RF: Right Front, LH: Left Hind, RH: Right Hind) is displayed with its corresponding robot contact state $\nnOut$.}
                \vspace{-0.5cm}
            \end{table}
            The dataset presented in \citep{lin2021deep_contact_estimation} is composed of output samples $\nnOut \in \R^{16}$, where each dimension of $\nnOut$ represents a logit of a specific contact state, among the $16$ different combinations of each of the $4$ legs possible binary contact states. The input samples $\nnOut=\{\vz_i\}_{i=0}^{150} \in \R^{54\times150}$, are a history of $150$ samples $\vz = [\qj, \dqj, {\bm{a}}, {\angvel}, {\vp}, {\vv}] \in \R^{54}$. Where $\qj \in \R^{\nj}, \dqj \in \R^{\nj}, {\bm{a}} \in \R^{3}, {\angvel} \in \R^{3}, {\vp} \in \R^{12}, {\vv}  \in \R^{12}$ are the MIT-Mini-Cheetah robot joint-space positions, velocities, base linear acceleration, base angular velocity, and each of the four legs feet's position and velocities, respectively, referenced to the robots base frame $\base$.
        
            The function approximator to learn is expected to be approximately equivariant to the reflection group $\Cyclic[2]$, considering the sagittal symmetry of the robot morphology. Therefore:
            \begin{align}
                \g \cdot \nnOut = \nn{\g \cdot \nnIn} \quad | \quad \g \in \G=\Cyclic[2] 
                \label{eq:contact_nn_def}
            \end{align}
        
        \subsubsection{Determination of the input and output representations $\rep[\nnInSpace]{\g}, \rep[\nnOutSpace]{\g} \;|\; \g \in \G$}
            \label{sec:sup_id_reps_contact}
                The MiniCheetah robot evolves in the Euclidean space of $3$-dimensions. Therefore its configuration space can be decoupled into $\ConfSpace \doteq \EG[3] \times \ConfSpaceJS$. After identifying their symmetry groups and their corresponding $\EG[3]$ and $\ConfSpaceJS$ representations ($\rep[{\EG[3]}]{\g}, \rep[{\ConfSpace}]{\g} \; |\;\forall\; \g \in \G$), we can identify the representations of the input and output spaces of the NN function approximator (\cref{eq:contact_nn_def}), considering that:
                \begin{multline}
                \small
                    \rep[\nnOutSpace]{\g}
                     \nnOut
                    \approx  \\
                    f\Biggl(
                         \ubcolor{awesomeblue}{
                            \begin{bsmallmatrix}
                                \rep[\ConfSpaceJS]{\g}& \bm{0}& \bm{0}& \bm{0}& \bm{0}& \bm{0}\\
                                \bm{0}& \rep[\ConfSpaceJS]{\g}& \bm{0}& \bm{0}& \bm{0}& \bm{0}\\
                                \bm{0}& \bm{0}& \rep[{\EG[3]}]{\impg}& \bm{0}& \bm{0}& \bm{0}\\
                                \bm{0}& \bm{0}& \bm{0}& \rep[{\EG[3]}]{\impg}& \bm{0}& \bm{0}\\
                                \bm{0}& \bm{0}& \bm{0}& \bm{0}& \rep[{\vp}]{\g}& \bm{0}\\
                                \bm{0}& \bm{0}& \bm{0}& \bm{0}& \bm{0}& \rep[{\vp}]{\g}\\
                            \end{bsmallmatrix}
                         }{\rep[\nnInSpace]{\g}}
                         \ubcolor{black}{
                            \begin{bsmallmatrix}
                                \qj\\ \dqj\\ {\bm{a}}\\ {\angvel}\\ {\vp}\\ {\vv}\\ 
                            \end{bsmallmatrix}
                        }{\nnIn}
                    \Biggl) \\
                    |\; \forall \; (\g,\,\impg) | \g \in \G, \impg \in \EG[3]
                    \nonumber
                \end{multline}
                Where the representation $\rep[\vp]{\g}$ acting on ${\vp} \in \R^{12}$ and ${\vv} \in \R^{12}$ is determined understanding that each of the feet positions $
                        (\vp{_{RF}}, 
                        \vp{_{LF}}, 
                        \vp{_{RH}}, 
                        \vp{_{LH}})$ 
                and velocities $(
                        \vv{_{RF}}, 
                        \vv{_{LF}}, 
                        \vv{_{RH}}, 
                        \vv{_{LH}})$
                are simply vectors living in $\EG[3]$. Thus, we must apply the euclidean action $\rep[{\EG[3]}]{\impg}$ while at the same time permuting the feets (similar to the permutation of the kinematic subchains described by $\rep[\ConfSpaceJS]{\g}$): 
                \begin{align}
                    \small 
                    \g \cdot \vp &= 
                        \ubcolor{awesomeblue}{
                            \rep[\R^{4}]{g} \otimes \rep[{\EG[3]}]{\impg} 
                        }{\rep[\vp]{\g}} \vp,  
                    \;\;|\; \forall (\g,\,\impg) | \g \in \G, \impg \in \EG[3]
                    \label{eq:feet_permutation}
                    \\
                    \nonumber
                    &= 
                    \begin{bsmallmatrix}
                        0 & 1 & 0 & 0 \\
                        1 & 0 & 0 & 0 \\
                        0 & 0 & 0 & 1 \\
                        0 & 0 & 1 & 0 \\
                    \end{bsmallmatrix} \otimes \rep[{\EG[3]}]{\impg} 
                    \begin{bsmallmatrix}
                        \vp{_{RF}}\\ 
                        \vp{_{LF}}\\ 
                        \vp{_{RH}}\\ 
                        \vp{_{LH}}\\ 
                    \end{bsmallmatrix} \;\;|\; \g \neq e
                    \\
                    \g \cdot \vv &= 
                        \ubcolor{awesomeblue}{
                            \rep[\R^{4}]{g} \otimes \rep[{\EG[3]}]{\impg} 
                        }{\rep[\vp]{\g}} \vv  
                    \;\; |\; \forall (\g,\,\impg) | \g \in \G, \impg \in \EG[3]
                    \label{eq:feet_permutation_c2}
                    \\
                    \nonumber
                    &=
                    \begin{bsmallmatrix}
                        0 & 1 & 0 & 0 \\
                        1 & 0 & 0 & 0 \\
                        0 & 0 & 0 & 1 \\
                        0 & 0 & 1 & 0 \\
                    \end{bsmallmatrix} \otimes \rep[{\EG[3]}]{\impg} 
                    \begin{bsmallmatrix}
                        \vv{_{RF}}\\ 
                        \vv{_{LF}}\\ 
                        \vv{_{RH}}\\ 
                        \vv{_{LH}}\\ 
                    \end{bsmallmatrix} \;\;|\; \g \neq e
                \end{align}
                Being $\rep[\R^{4}]{\g} \;|\;\forall \g \in \Cyclic[2]$ two regular representations of $\Cyclic[2]$ stack in block diagonal form to form a permutation representation in $4$-dimensional space, representing the right-left symmetries of the $4$ kinematic tree's subchains (four legs). The nature of the representation $\rep[\ConfSpaceJS]{\g} \doteq \rep[\R^{4}]{\g} \otimes \Identity[n_s]$ might be better understood if you consider that we apply $\rep[\R^{4}]{\g}$ to each set of symmetric DoF in the kinematic three, which for the Mini-Cheetah is $n_s = 3$ sets of symmetric DoF (see \cref{fig:sup_solo_symmetry_planes}-right). See simpler examples in \cref{fig:examples_morphological_symmetries}.
            
                Lastly, the representation for the contact state $\rep[\nnOutSpace]{g}$ is given by the permutation matrix relating $\nnOut$ and $\g\cdot \nnOut$ described in \cref{table:contact_state_symmetry}.
            
        \subsubsection{Details on dataset partitioning}
        \label{sec:sup_contact_dataset}
            \begin{table}[t]
                \label{table:hps}
                \begin{center}
                \begin{tabular}{ c | c | c c c }
                    \textbf{Robot} & \textbf{Model} & $\G$ & \textbf{lr} & \textbf{Samples}\\
                    \hline
                    Solo & MLP/EMLP & $\Cyclic[2]\;\&\; \KleinFour$ & $2.4\times10^{-3}$ & 100k \\ 
                    Atlas & MLP/EMLP & $\Cyclic[2]$ & $1.5\times10^{-3}$ & 500k\\  
                    \hline
                    MiniCheetah & CNN & $\Cyclic[2]$ & $1.0\times10^{-4}$ & \multirow{2}{2em}{730k}\\
                    MiniCheetah & E-CNN & $\Cyclic[2]$ & $1.0\times10^{-5}$ & 
                \end{tabular}
                \end{center}
                \caption{Robot, Models and Dataset parameters.}
                \vspace{-0.5cm}
            \end{table}
            We modified the original dataset partitioning to properly evaluate the generalization capacity of the models. The original dataset was composed of 15 different recordings varying ground type and gait type used during data collection (most recordings were performed on a 
             trot gait, which heavily biased the dataset to contact states $0$, $6$, and $9$ of \cref{table:contact_state_symmetry}). 
             
             The authors of \citep{lin2021deep_contact_estimation} partitioned all 15 recordings into ($70\%, 15\%, 15\%$) training, validation and testing. This partition was made such that the first $70\%$ time-samples of each recording were assigned for training, the following $15\%$ to validation, and the rest for testing. 
             
             Because we are interested in studying the generalization capacity of the models and the out-of-training-distribution performance, we modified this partitioning such that among the $15$ different recordings we selected randomly $5$ recordings for testing, and the remaining $10$ recordings were used for training splitting these recordings into ($85\%$, $15\%$) training and validation splits as in \cite{lin2021deep_contact_estimation}, that is, for each recording, the first $85\%$ data-samples go for training and the remaining for validation.  
            
            The selected training-validation recordings were: air walking gait,  concrete difficult slippery,  concrete left circle,  middle pebble, rock road, asphalt road, concrete galloping, grass, old asphalt road, sidewalk. While the selected testing recordings were: air jumping gait,  concrete pronking,  concrete right circle,  forest,  small pebble.
    
        \subsection{Mitigation of suboptimal asymmetries in model performance}
        \label{sec:sup_contact_asymmetries}
            When comparing individual leg classification we see that the equivariant model converges to having a similar performance for each symmetric pair of legs, while the unconstrained models converge to an asymmetrical suboptimal state favoring the contact detection of one leg at the expense of reduced performance for the symmetric leg (see LF and RF f1-scores). This asymmetrical performance is attributed to the CNN and CNN-aug models learning to extract temporal features for both symmetric legs separately, increasing the likelihood of converging to asymmetrical local minima. On the contrary, the equivariant model E-CNN can be thought of as learning to extract a single set of \textit{symmetric} temporal features for each symmetric pair of states (a consequence of the model's equivariance and parameter sharing). This implies that the temporal features used for determining the contact state of, say the left frontal leg, would also be used to determine the contact state of the symmetric leg, the right frontal leg, and vice-versa.
    
        \subsection{Equivariant Conv1D layers}
        For details on the construction of the Equivariant 1D Convolutional layers reefer to \footref{foot:code}. Note that the symmetry of a single time-sample $\vz_i$ is shared across all time-samples $\nnOut=\{\vz_i\}_{i=0}^{150}$.     
        \begin{figure*}[!t]
            \centering
            \begin{coloredFrame}[boxsep=0pt,left=4pt,right=4pt,top=15pt,bottom=15pt]{awesomeblue}{}
                \setlength{\columnseprule}{1pt}
                \def\columnseprulecolor{\color{white}}
                \setlength\columnsep{20pt}
                \begin{multicols}{2}    
                    \begin{center}
                        \textbf{Tri-Finger Robot} $\G=\Cyclic[3]$
                        \includegraphics[height=4cm]{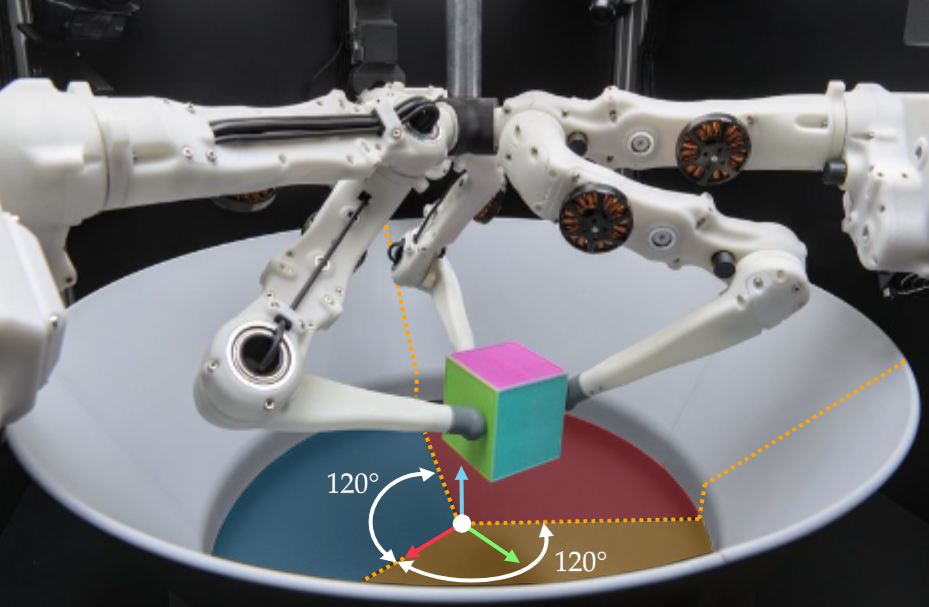} 
                    \end{center}
                    This fixed-based robot is symmetric w.r.t. rotations of space by $\theta=120^{\circ}$ in the vertical axis. Therefore, its symmetry group is the cyclic group of order three ($\G=\Cyclic[3]$). To identify this symmetry group we apply the procedure in \cref{sec:identification_dms_g}:
                    \begin{enumerate}
                        \item \textbf{Identify $\baseSE$ and $\Inertia[\base]$}: As a fix-based robot, we define $\baseSE$ to be the mounting structure supporting each finger, and the gray disk delimiting the workspace (see image). 
                        \item \textbf{Identify symmetries of $\Inertia[\base]$}:
                        The inertia of this virtual base $\Inertia[\base]$ is invariant to rotations by $120^{\circ}$ in the vertical axis. I.e., $\Inertia[\base]$ is invariant to $\baseSE \rep[{\EG[3]}]{\impg}^{\text{-}1} | \impg \in \{e, \impg[\theta], \impg[\theta]^2\} \equiv \Cyclic[3]$ (\cref{eq:diff_g_imp_g}). 
                        \item \textbf{Identify modularity in the kinematic tree}: There are three symmetric kinematic subchains. Each finger is composed of replicated versions of the same bodies. 
                        \item \textbf{Identify the DMS group $\G$}:  
                    \end{enumerate}
                    Consider that the transformation $\rep[{\EG[3]}]{\g}\baseSE \doteq \rep[{\EG[3]}]{\impg}\baseSE \rep[{\EG[3]}]{\impg}^{\text{-}1}$ (\cref{eq:diff_g_imp_g}) can be interpreted as a rotation of the virtual base by $\theta^\circ$ followed by a rotation $-\theta^\circ$ in the $z$ axis. Thus respecting the fix-base constraint of the system. Denote the joint-space $\q = \qj = [\q[f1]^\transpose, \q[f2]^\transpose, \q[f3]^\transpose]^\transpose$ be composed of each finger's DoF ($\q[fi] \in \R^{3}$).
                    \\
                    
                    Then we can define $\rep[\ConfSpaceJS]{g} \doteq \rep[\R^3]{\impg} \otimes \Identity[3] \;|\; \impg \in \Cyclic[3]$. Being $\rep[\R^3]{\cdot}$ the permutation representation of $3$ elements of $\Cyclic[3]$ ($3$ kinematic subchains). For the generator action of the group this is $\rep[\R^3]{\impg} = \begin{bsmallmatrix}
                        0 & 1 & 0\\
                        0 & 0 & 1\\
                        1 & 0 & 0\\
                    \end{bsmallmatrix}$. \\
    
                    Lastly, we verify if $\G=\Cyclic[3]$ by testing all tentative group actions for DMSs \cref{eq:discrete_morphological_symmetries}. \\
    
                    \textbf{Augmentation of data samples}: Say we collect a dataset of robot states $(\q,\dq)$ and cube states $\memSE[C]$ at every time step $t$, to train the manipulation policy \citep{funk2021trifinger}. To obtain the symmetric states, at every $t$, we need to understand that since we are imitating the effect of a true rotation of space $\impg$, the symmetric states are obtained by ($\gq, \gdq$) and ($\impg \cdot \memSE[C] \doteq \rep[{\EG[3]}]{\impg}\memSE[C]$). 
                    
                    \vfill\null
                    \columnbreak
                    
                    \begin{center}
                        \textbf{Bolt Bipedal Robot $\G=\Cyclic[2]$}
                        \href{https://imgur.com/a/CYj8wTQ}{\includegraphics[height=4cm]{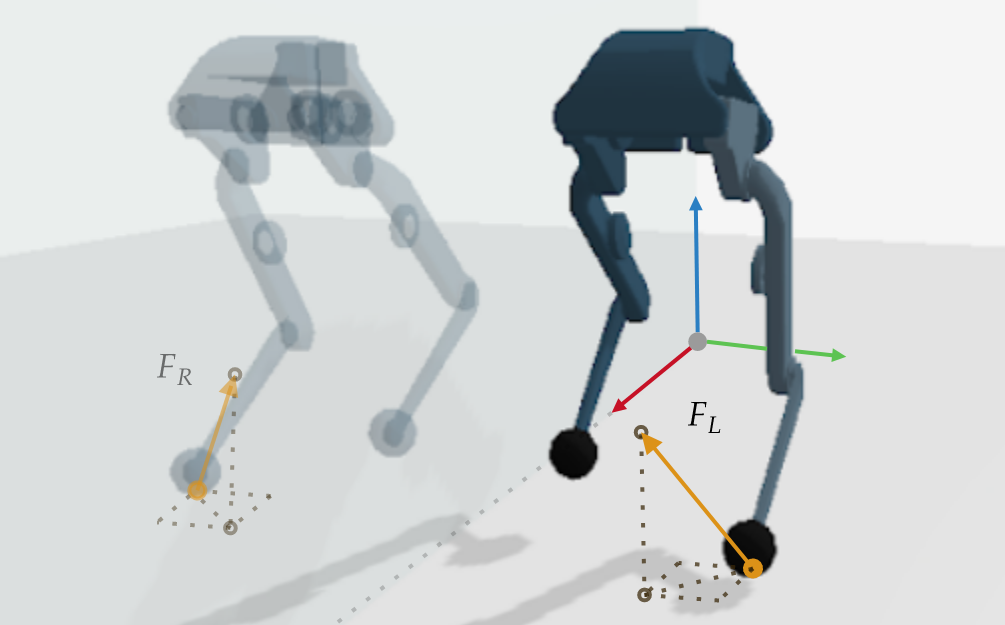}}
                    \end{center}
                    Bolt is a bipedal robot with a sagittal plane reflection symmetry ($\G=\Cyclic[2]$). This morphological symmetry allows it to imitate the effect of arbitrary reflections of space ($\impg \in \EG[3]$) by re-configuring its base and legs. To identify this symmetry group we apply the procedure in \cref{sec:identification_dms_g}:
    
                    \begin{enumerate}
                        \item \textbf{Identify $\baseSE$ and $\Inertia[\base]$}:
                        $\baseSE$ is the robot base (hips) body, with its corresponding inertia $\Inertia[\base]$
                        \item \textbf{Identify symmetries of $\Inertia[\base]$}:
                        The base body has symmetrical mass distribution w.r.t the sagittal plane. Thus, $\Inertia[\base]$ is invariant to the transformation $\baseSE \rep[{\EG[3]}]{\impg}^{\text{-}1} | \impg \in \{e, \impg[s]\} \equiv \Cyclic[2]$ (\cref{eq:diff_g_imp_g}). 
                        \item \textbf{Identify modularity in the kinematic tree}: There are two symmetric kinematic subchains. The left leg subchain and bodies are reflected versions of the right leg subchain and bodies. 
                        \item \textbf{Identify the DMS group $\G$}:  
                    \end{enumerate}
                    Since a reflection w.r.t to the sagittal plane would imply a true reflection of the rigid bodies of the legs, we need to \textit{permute} each body in the kinematic tree with each reflected version. Denote the joint-space $\qj=[\q[L]^\transpose, \q[R]^\transpose]^\transpose$ as composed of the left $L$ and right $R$ legs' DoF ($\q[L/R] \in \R^3$).
                    Denote the sign-relation between the DoF of the Left and right legs' degrees of freedom as $\vs_{L|R} \in \R^3$. \\ 
    
                    Then we can define $\rep[\ConfSpaceJS]{g} \doteq \rep[\R^2]{\impg} \otimes (\vs_{L|R}\Identity[3]) \;|\; \impg \in \Cyclic[2]$. Being $\rep[\R^2]{\cdot}$ the permutation representation of a $2$ elements of $\Cyclic[2]$($2$ kinematic subchains). For the non-trivial action of the group this is $\rep[\R^2]{\impg[s]} = \begin{bsmallmatrix}
                        0 & 1\\
                        1 & 0
                    \end{bsmallmatrix}$. \\
    
                    Lastly, recalling the definition of $\rep[{\EG[3]}]{\g}$ in \cref{eq:diff_g_imp_g}, we verify if $\G=\Cyclic[2]$ by testing all tentative group actions for DMSs \cref{eq:discrete_morphological_symmetries}. 
                    \\
    
                    \textbf{Augmentation of data samples}: Say we collect a dataset of robot states $(\q,\dq)$ and ground reaction forces $(\vf_L, \vf_R)$, that we transform to the space of generalized forces as $(\genForces_{f_L},\genForces_{f_R})$, at every timestep $t$. This dataset can be used to train a reactive locomotion policy as in \citet{ordonez2022adaptable}. The symmetric states, at every $t$, are then defined as: $(\gq,\gdq)$ and $(\g \cdot \genForces_{f_L},\g \cdot\genForces_{f_R}) \equiv (\rep[\ConfSpace]{\g} \genForces_{f_L},\rep[\ConfSpace]{\g}\genForces_{f_R})$
            \end{multicols}
            \end{coloredFrame}
            \caption{\textbf{Tutorial} example morphological symmetries of the Tri-Finger~\citep{funk2021trifinger} and Bolt robots.
            }
            \label{fig:examples_morphological_symmetries}
        \end{figure*}
         \begin{figure*}[!t] 
                \centering
                \includegraphics[width=.9\textwidth]{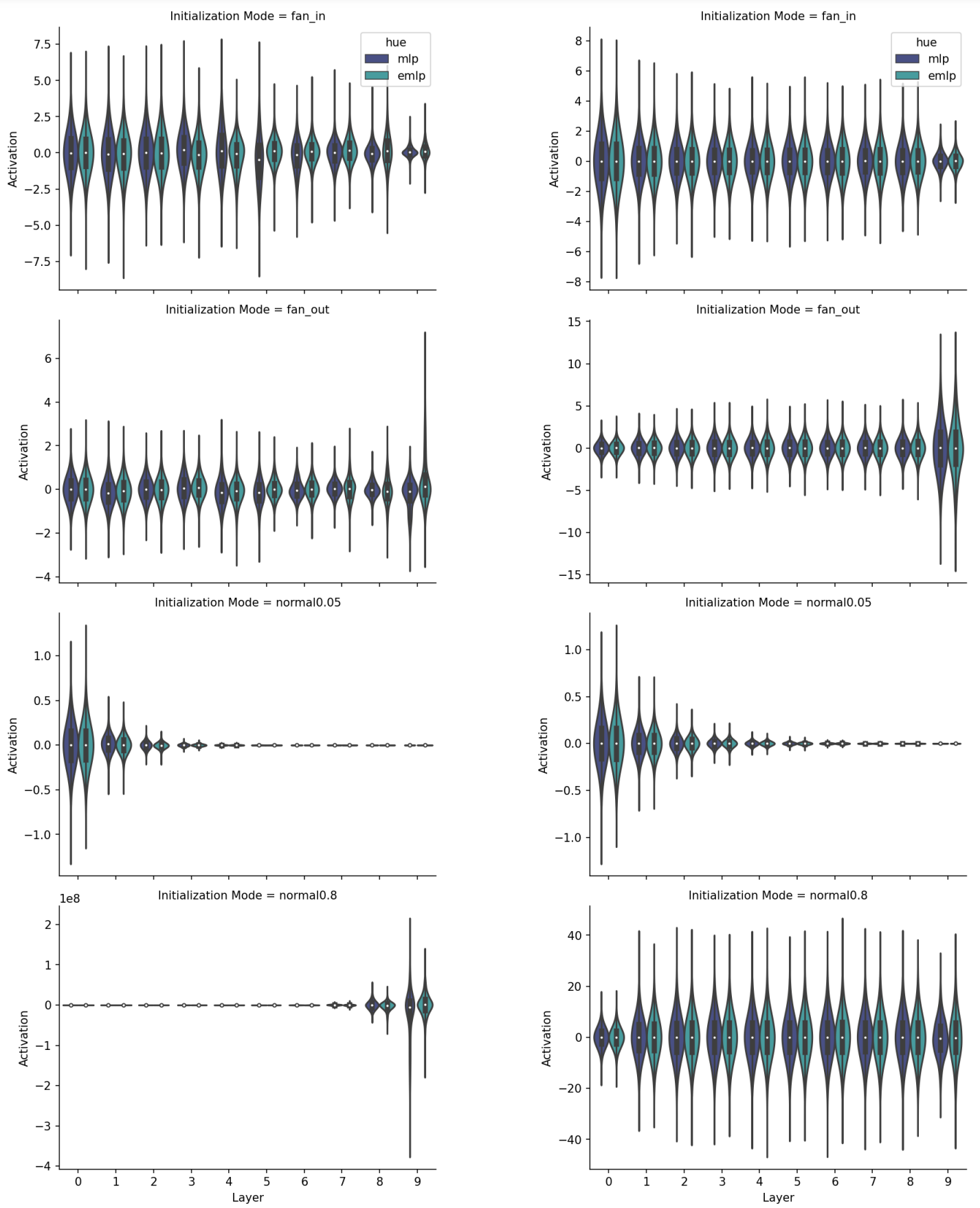}
                \label{fig:sup_init_tanh}
                \caption{
                Comparison of the initialization method of unconstrained layers of \citet{he2015keiming_delving} with our initialization method for equivariant layers. Left and right columns correspond to MLP \& EMLP architectures with $\sigma=ReLu$ (left) and $\sigma=Tanh$ (right) non-linearities. Each row shows different initialization methods differing in the variance of the initialization distribution of the layer's trainable parameters. First and second rows show the forward and backward propagation cases of \cite{he2015keiming_delving} for MLP and of \cref{sec:sup_initialization_equivariant_layers} for EMLP, with $\Var(^{l}\basisCoefEV) \doteq \sfrac{m}{(\lambda_{^{l}\basisEquivT}\lambda_{\nnAct})}$ and $\Var(^{l}\basisCoefEV) \doteq \sfrac{n}{(\lambda_{^{l}\basisEquivT}\lambda_{\nnAct})}$, respectively. In these cases, the variance of activations through the network depth remains nearly constant, as desired. The last two rows show the initialization of layer parameters with a constant variance of $0.05^2$ and $0.8^2$, illustrating scenarios of activations vanishings and exploiting. All intermediate layers have 256 neurons. In the equivariant case, the network is $\KleinFour$-equivariant.
                }
                \label{fig:sup_initialization}
            \end{figure*}
    
    \section{Limitations}
    \label{sec:sup_limitations}
    
        
        Our work makes two main assumptions:
        \begin{enumerate}
            \item \textbf{Symmetries are exact}: By assuming that a dynamical system has exact and not approximate symmetries we are departing from the real-world nature of DMSs since for any robotic system in the real-world the manufacturing and assembly process introduces errors/tolerances in the kinematic and dynamic parameters of each of the robot's bodies. Likewise, the dynamics of animals in nature are not perfectly equivariant since morphological symmetries are only approximate symmetries. Although exact symmetries seem to be a strong assumption, in practice, the reality is that it is a common assumption in the fields of robotics and control theory, in which idealized models of the dynamics are often assumed (in simulation and real-world). 
            
            On \cref{sec:experiments} we show that the exact symmetry bias is justifiable and beneficial for learning function approximators processing the dynamics of  approximately symmetrical systems in the real world. However, the authors highlight the necessity to properly address the case of approximate equivariance, which we leave to future work. To address this case, system identification techniques \cite{simpkins2012systemID} have been wildly used to approximate the deviation of the kinematic and dynamic parameters from the assumed values. While in the case of $\G$-equivariant NN     
            \citet{wang2022approximately, finzi2021residualPP} provide clear and valuable approaches to learn approximate $\G$-equivariant NN.
        
            It is relevant to highlight that, in physics-based simulation, the most common practice is to work with the idealized model of dynamics. Thus, the assumption of exact symmetries is justifiable and encouraged in applications where simulation is a relevant tool.
            
            \item \textbf{Symmetry group is finite}: This work narrows its focus to DMSs, even do the definition of DMS \cref{eq:discrete_morphological_symmetries} can be relaxed to cover continuous morphological symmetries (an example of this continuous symmetry is the capability of a robot arm manipulator to imitate rotations of space by rotating its first degree of freedom). This assumption allows us to compute in linear time the basis $\basisEquivT$ of equivariant linear maps (restricted to groups with finite regular matrix representations).
        \end{enumerate}
    
}

\end{document}